\pdfoutput=1
\documentclass[10pt,twocolumn,letterpaper]{article}

\newif\ifarxiv
\arxivtrue

\ifarxiv
\usepackage[pagenumbers]{cvpr} %
\else
\usepackage{cvpr}              %
\fi 

\def\MYTITLE{Iterative Event-based Motion Segmentation by\\ Variational Contrast Maximization}

\usepackage{amsmath,amssymb,amsfonts}
\usepackage{graphicx}
\usepackage{booktabs}
\usepackage{color}
\definecolor{cvprblue}{rgb}{0.21,0.49,0.74}
\ifarxiv
\usepackage[breaklinks,colorlinks,allcolors=cvprblue,bookmarks=true,bookmarksnumbered=true]{hyperref}
\hypersetup{
  pdftitle={\MYTITLE},
  pdfsubject={Computer Vision, Robotics},
  pdfauthor={Shintaro Shiba, Yoshimitsu Aoki, Guillermo Gallego},
  pdfkeywords={Event Cameras, Segmentation, Motion Estimation}
}
\usepackage[absolute]{textpos}
\else
\usepackage[pagebackref,breaklinks,colorlinks,allcolors=cvprblue]{hyperref}
\fi

\usepackage{siunitx}
\usepackage{makecell}
\usepackage{adjustbox}

\usepackage{algorithm}
\usepackage{algorithmic}

\usepackage{cuted}
\def\tref{t_\text{ref}} %
\def\pol{p} %

\def\cE{\mathcal{E}} %
\def\numEvents{N_e} %
\def\numPixels{N_p} %
\def\Warp{\mathbf{W}}

\def\bx{\mathbf{x}}
\def\bparams{\boldsymbol{\theta}}

\def\pol{p}
\def\velflow{\mathbf{v}}
\def\linvel{\mathbf{V}} %
\def\angvel{\boldsymbol{\omega}} %
\def\bmu{\boldsymbol{\mu}}

\def\IWE{I}
\def\depth{Z} %

\def\IWE{\mathrm{IWE}} %

\def\costFunc{\mathcal{L}} %

\def\cN{\mathcal{N}} %

\def\bzero{\mathbf{0}}

\newcommand{\unum}[2]{\multicolumn{1}{c}{\underline{\tablenum[table-format={#1}]{#2}}}}  %
\newcommand{\bnum}[1]{\bfseries #1}

\newcommand{\novalue}{{\textendash}}

\definecolor{light-gray}{gray}{0.6}
\newcommand\gframe[1]{{\color{light-gray}\frame{#1}}}

\begin{document}

\ifarxiv
\definecolor{somegray}{gray}{0.5}
\newcommand{\darkgrayed}[1]{\textcolor{somegray}{#1}}
\begin{textblock}{11.5}(2.25, 0.8)  %
\begin{center}
\darkgrayed{This paper has been accepted for publication at the\\
IEEE Conference on Computer Vision and Pattern Recognition (CVPR) Workshops, Nashville, 2025.
\copyright IEEE}
\end{center}
\end{textblock}
\fi

\title{\MYTITLE}

\author{Ryo Yamaki$^{1}$, Shintaro Shiba$^{1,2}$, Guillermo Gallego$^{2,3}$, and Yoshimitsu Aoki$^{1}$\\
$^{1}$~Keio University, Japan.
$^{2}$~Technische Universit\"at Berlin, 
$^{3}$~Einstein Center Digital Future, \\Robotics Institute Germany, and Science of Intelligence Excellence Cluster, Germany.
\vspace{-1.8ex}
}

\maketitle

\begin{strip}
\centering
\vspace{-2ex}
\includegraphics[width=0.9\linewidth]{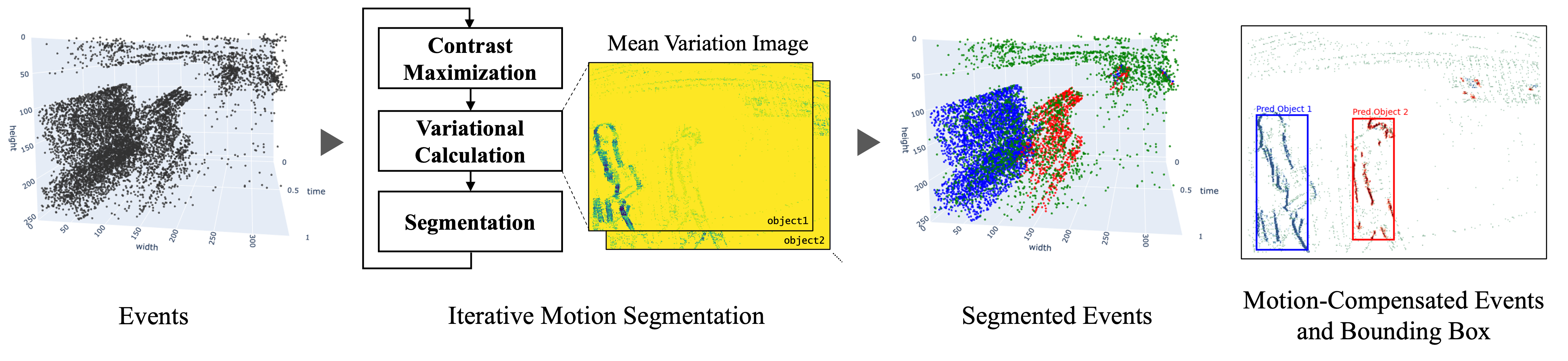}
\captionof{figure}{\emph{Overview}. The proposed method relies on event data, and achieves motion segmentation by iteratively running Contrast Maximization with respect to the parameters \emph{and} with respect to the event data.
The variation can be visualized as Mean Variation Image.
As results, we obtain the segmented event stream and motion-compensated images (i.e., images of warped events).
The scene (from \cite{Zhou21tnnls}) shows two pedestrians walking in a corridor while the camera slightly moves.
}
\label{fig:overview}
\end{strip}

\begin{abstract}
Event cameras provide rich signals that are suitable for motion estimation since they respond to changes in the scene.
As any visual changes in the scene produce event data,
it is paramount to classify the data into different motions (i.e., \emph{motion segmentation}), 
which is useful for various tasks such as object detection and visual servoing.
We propose an iterative motion segmentation method,
by classifying events into background (e.g., dominant motion hypothesis) and foreground (independent motion residuals),
thus extending the Contrast Maximization framework.
Experimental results demonstrate that the proposed method successfully classifies event clusters
both for public and self-recorded datasets,
producing sharp, motion-compensated edge-like images.
The proposed method achieves state-of-the-art accuracy on moving object detection benchmarks with an improvement of over 30\%,
and demonstrates its possibility of applying to more complex and noisy real-world scenes.
We hope this work broadens the sensitivity of Contrast Maximization with respect to both motion parameters and input events, 
thus contributing to theoretical advancements in event-based motion segmentation estimation.
\url{https://github.com/aoki-media-lab/event_based_segmentation_vcmax}
\end{abstract}

\section{Introduction}
\label{sec:intro}

\begin{table*}[t!]
\centering
\adjustbox{max width=\linewidth}{%
\setlength{\tabcolsep}{3pt}
\begin{tabular}{V{\linewidth}V{\linewidth}V{\linewidth}V{\linewidth}V{\linewidth}}
\toprule 
 & \textbf{Mitrokhin et al.} \cite{Mitrokhin18iros} & \textbf{EMSMC} \cite{Stoffregen19iccv} & \textbf{EMSGC} \cite{Zhou21tnnls} & \textbf{This work}\tabularnewline
\midrule 
\textbf{Problem unknowns} & Motion parameters $\{\theta_{j}\}_{j=1}^{N_{c}}$,

per-pixel cluster labels

$\{\ell_{\mathbf{x}}\},$ $\ell_{\mathbf{x}}\in\{1,...,N_{c}\}$ & Motion parameters $\{\theta_{j}\}_{j=1}^{N_{c}}$,

per-event cluster

probabilities $\{p_{kj}\}$ & Motion parameters $\{\theta_{j}\}_{j=1}^{N_{c}}$,

per-event cluster labels

$\{\ell_{k}\},$ $\ell_{k}\in\{1,...,N_{c}\}$ & Motion clusters $\{\theta_{j},\mathcal{E}_{j}\}_{j=1}^{N_{c}}$

(same meaning as EMSGC)\tabularnewline
\midrule 
\textbf{Number of clusters $N_{c}$} & Automatically computed (by 

detection hyper-parameters). & Given & Automatically computed

(by means of MDL loss term) & Automatically computed\tabularnewline

\midrule 
\textbf{Objective function} & Event alignment ($\approx$CMax) & Motion compensation (CMax) & Motion compensation (CMax)

and two priors (Potts \& MDL) & Motion compensation

(CMax)
\tabularnewline
\midrule 
\textbf{Estimation approach} & Greedy (dominant motion

and residuals) & Joint, by Expectation-

Maximization
& Joint, by alternating solvers 

on a Markov Random Field & Greedy (dominant motion

and residuals)\tabularnewline
\midrule 
\textbf{Optimization method} & Gradient descent and

thredholding & Conjugate gradient and 

``closed-form'' formula & Conjugate gradient and

graph cut
& Conjugate gradient and 

thresholding
\tabularnewline

\midrule 
\textbf{Motion estimation} & Motion compensation on

residual events
& CMax with probabilities $p_{kj}$ 

(using weighted IWEs) & CMax using selected events 

per cluster & CMax on residual events\tabularnewline
\midrule 
\textbf{Event--cluster}

\textbf{associations} & Hard (0 or 1, dominant

motion or residuals) & Soft (probabilities $p_{kj}$) & Hard $\ell_{k}\in\{1,...,N_{c}\}$ & Hard (0 or 1, dominant

motion or residuals)
\tabularnewline
\midrule 
\textbf{Segmentation }

\textbf{(Classification)} & Per-pixel, by thresholding 

time-surface dispersion. & Per-event, closed-form partition-

of-unity formula based on

local contrast & Per-event, using a graph

through the events and

graph cut(s) & Per-event, by thresholding 

the loss variation\tabularnewline
\bottomrule
\end{tabular}
}
\caption{\label{tab:methodComparison}Comparison of event-based motion segmentation methods based on the CMax framework.}
\end{table*}

Event cameras are bio-inspired vision sensors that provide asynchronous responses to the motion of edges
with high dynamic range (HDR) and minimal blur at high temporal ($\si{\micro\second}$) resolution \cite{Posch14ieee}.
The signals from event cameras are naturally suitable for motion estimation, especially in challenging real-world scenarios for conventional (frame-based) cameras.
Segmenting different scene motions in image space is a paramount task for various downstream tasks, such as object detection and visual servoing.

Motion segmentation for event cameras classifies each event into different clusters that represent the motions of the objects causing the events.
However, 
since event cameras produce a high-dimensional output stream (``events'') due to their asynchronous working principle,
finding different motion clusters in the event stream is a challenging task.
Conventional frame-based methods cannot be easily adapted to event data as the data types are fundamentally different \cite{Gallego20pami}.
In some previous event-based motion segmentation methods,
the number of clusters is predetermined. For example, \cite{Stoffregen19iccv} proposed a method based on the Contrast Maximization (CMax) framework \cite{Gallego18cvpr} to simultaneously optimize two sets of unknowns: 
the motion parameters of each cluster and the probability that each event belongs to each cluster.
However,
providing the number of clusters for a new scene is not straightforward.
Hence, it is paramount to develop methods that iteratively segment motion clusters by classifying events into different clusters.

In this work, we propose a novel algorithm that classifies events
iteratively to achieve motion segmentation, extending the CMax framework (\cref{fig:overview}).
The proposed variational approach defines the scores about how much each event aligns with the current motion hypothesis after estimating the dominant motion of the scene, 
and it runs iteratively on the events that do not conform to the dominant motion (i.e., rest of events or ``residual events'').
Experimental results demonstrate that the proposed method successfully classifies event clusters both for simple datasets and public datasets, producing sharp, motion-compensated images.
Also, the proposed method is useful for object detection, where it achieves the state-of-the-art bounding box accuracy by $30\%$ improvement.

Our contributions can be summarized as follows:
\begin{enumerate}
    \item We propose a novel method to classify events with a given motion hypothesis, based on the strength of the first variation of the CMax loss function with respect to the events (\cref{sec:method:variation}).
    \item We propose an iterative motion segmentation with the above classification, by estimating dominant motion of the current events and the residual events that do not support the dominant motion (\cref{sec:method}).
    \item We conduct comprehensive experiments on four different datasets, evaluating on simple scenes and real-world data, showcasing the efficacy of the proposed method on both motion segmentation and object detection (\cref{sec:experiment}). 
\end{enumerate}

\section{Related Work}
\label{sec:relatedworks}

Event-based motion segmentation is a task that
classifies events into different motion clusters.
Prior work can be categorized into model-based and learning-based approaches.
Model-based approaches cluster the input events (spatio-temporal stream)
that best aligns with different motions.
Extending the CMax framework \cite{Gallego17ral,Gallego18cvpr,Gallego19cvpr}, 
Stoffregen et al. \cite{Stoffregen19iccv} propose
``Event-based Motion Segmentation by Motion Compensation'' (EMSMC), 
an Expectation-Maximization (EM) algorithm that estimates the motion parameters and the event-cluster classification probabilities at the same time, 
maximizing the contrast of a compound image of warped events (IWE).
In \cite{Parameshwara21icra}, motion compensation is combined with feature tracking,
repeatedly fitting and merging models with a scene segmented into multiple motions.
Zhou et al. \cite{Zhou21tnnls} propose an optimization method using a graph-cut technique for clustering events. 
They also propose a new dataset for the task.

Learning-based methods predict independently-moving object (IMO) masks, or alternatively predict other quantities (e.g., optical flow, scene depth, etc.) and estimate IMOs from them \cite{Alkendi24tim,zhou2024jstr,Wang24evmoseg,Arja24arxiv,Alkendi25tmm}.
A typical approach uses optical flow networks before clustering IMOs, such as unsupervised-learning--based method \cite{Wang24evmoseg}.
Other scene quantities, such as semantic labels \cite{Jiang24visapp} and the scene depth and camera pose \cite{Georgoulis24threedv}, can also be utilized.
Works like \cite{Mitrokhin19iros,Burner22evimo2} follow this approach and release the de-facto standard dataset (EVIMO).
Recently, the spatio-temporal characteristics of event data are further investigated, by exploring model architectures, such as 
graph convolutional network \cite{Mitrokhin20cvpr}
and spiking neural networks \cite{Parameshwara21iros}.
Additionally, Zhou et al. \cite{zhou2024jstr} propose a spatio-temporal joint inference method, focusing on the cylindrical shape formed by IMO in the space-time domain of the image plane. 

Our optimization-based method is most related to EMSMC \cite{Stoffregen19iccv}, where the expectation-maximization method is combined with the CMax framework. 
However, we propose an iterative approach that mitigates the dependency on the number of clusters and the initial values for the optimization, resulting in state-of-the-art accuracy in both motion segmentation and object detection.
\Cref{tab:methodComparison} summarizes the characteristics of the closely related but different approaches.
A more technical comparison is presented at the end of \cref{sec:method:variation}, after explaining our approach.

\begin{figure*}[t]
    \centering
    \includegraphics[width=\linewidth]{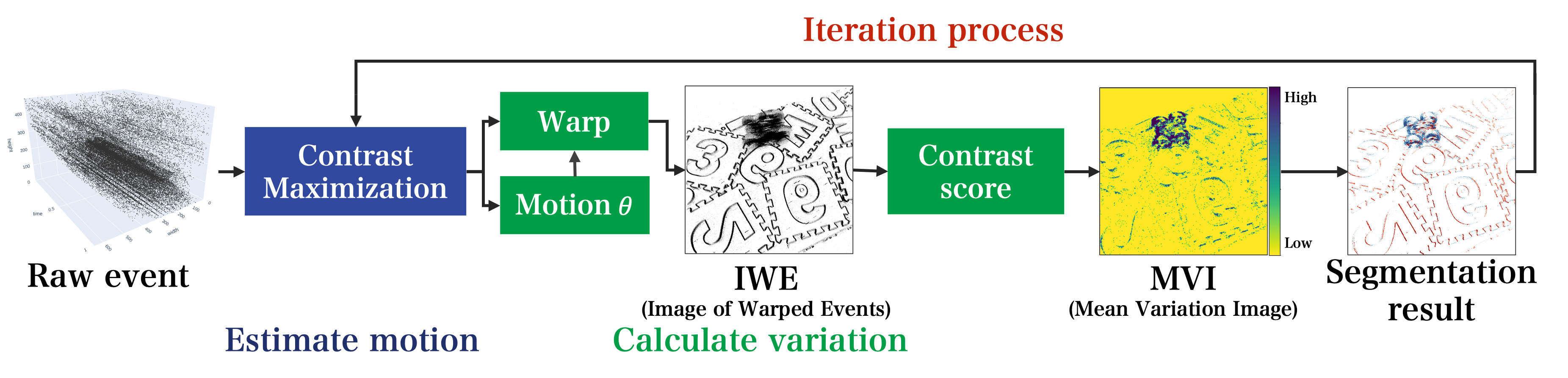}
    \caption{\emph{Block diagram of the proposed method}. 
    First, we use Contrast Maximization to estimate the dominant motion on the scene. 
    Using the estimated motion $\bparams$, we calculate the first variation of the contrast function with respect to the event coordinates \eqref{eq:variation}.
    The magnitude of the first variation can be visualized as a heat map \eqref{eq:gradientMap} (here, from yellow to blue): the higher it is, the greater the likelihood that the corresponding events do not conform to the estimated motion, i.e., they belong to an independent moving object (IMO).
    Thresholding the first variation classifies events into the ``fit'' events and ``residual'' events in terms of the currently estimated motion. 
    The ``fit'' events are removed (defining a segmented object or ``cluster'') and 
    the above steps are repeated on the residual events, until the final segmentation. 
    }
\label{fig:method}
\end{figure*}

\section{Method}
\label{sec:method}

In this section, first, we revisit the principles of event cameras (\cref{sec:method:eventcamera}).
Then, we cover the CMax framework (\cref{sec:contrast_maximization}), which is the basis of the proposed method.
The variational approach and motion segmentation method are explained in \cref{sec:method:variation}.

\subsection{Event Cameras}
\label{sec:method:eventcamera}

Instead of acquiring brightness images at fixed time intervals (e.g., frames),
event cameras record brightness changes asynchronously, called ``events'' \cite{Lichtsteiner08ssc,Gallego20pami}.
An event $e_k \doteq (\bx_k, t_k, \pol_{k})$ is triggered as soon as the logarithmic brightness at the pixel $\bx_k\doteq (x_k, y_k)^{\top}$ exceeds a preset threshold (i.e., contrast sensitivity) $C$.
Here, $t_k$ is the timestamp of the event with \si{\micro\second} resolution,
and polarity $\pol_{k} \in \{+1,-1\}$ is the sign of the brightness change:
\begin{equation}
\label{eq:event}
    L\left(\bx_k, t_k\right)-L\left(\bx_k, t_k-\Delta t_k\right)=\pol_k C.
\end{equation}

\subsection{Contrast Maximization}
\label{sec:contrast_maximization}

The CMax framework \cite{Gallego17ral,Gallego18cvpr,Gallego19cvpr} is powerful for various motion-estimation tasks, such as
rotational motion \cite{Gallego17ral,Kim21ral,Gu21iccv,Guo24tro,Shiba22aisy},
homographic motion \cite{Gallego18cvpr,Nunes21pami,Peng21pami},
optical flow \cite{Shiba22eccv,Paredes23iccv,Paredes24scirob,Shiba24pami,Hamann24eccv,Guo25e2fai},
and motion segmentation \cite{Mitrokhin18iros,Stoffregen19iccv,Zhou21tnnls,Parameshwara21icra}.
The first step in CMax is to transform
input events $\cE = \{e_k\}_{k=1}^{\numEvents}$ to warped events $\cE' = \{e'_k\}_{k=1}^{\numEvents}$ using a motion model $\Warp$,
\begin{equation}
\label{eq:warp}
e_k \doteq (\bx_k,t_k,\pol_k) \quad\stackrel{\Warp}{\mapsto}\quad
e'_k \doteq (\bx'_k,\tref,\pol_k).
\end{equation}
The warp $\bx'_k = \Warp(\bx_k,t_k; \bparams)$ transports each event along the point trajectory that passes through it,
until the reference time $\tref$ is reached.
The point trajectories are parametrized by $\bparams$, which is based on the motion model for estimation.
Then, an objective function (e.g., image contrast) \cite{Gallego19cvpr,Stoffregen19cvpr} measures the alignment of the warped events $\cE'$.
Many objective functions are defined in terms of the histogram of the warped events, or the image of warped events (IWE):
\begin{equation}
\label{eq:IWE}
\IWE(\bx;\bparams) \doteq \sum_{k=1}^{\numEvents} b_k \,\delta (\bx - \bx'_k(\bparams)).
\end{equation}
Each IWE pixel $\bx$ sums the values of the warped events $\bx'_k$ that fall within it.
$b_k=\pol_k$ if polarity is used or $b_k=1$ if polarity is not used.
The Dirac delta $\delta$ is in practice replaced by a smooth approximation~\cite{Ng22ral}, such as a Gaussian, $\delta(\bx-\bmu)\approx\cN(\bx;\bmu,\epsilon^2)$ with $\epsilon=1$~pixel.
A popular objective function $\costFunc(\bparams)$ is the variance of the IWE~\eqref{eq:IWE}:
\begin{equation}
\label{eq:IWEVariance}
\costFunc(\bparams) \equiv 
\frac{1}{|\Omega|} \int_{\Omega} (\IWE(\bx;\bparams)-\mu)^2 d\bx,
\end{equation}
with mean $\mu \doteq \frac{1}{|\Omega|} \int_{\Omega} \IWE(\bx;\bparams) d\bx$ over the image domain $\Omega$.
Hence, the alignment of the transformed events $\cE'$ (i.e., the candidate ``corresponding events'', triggered by the same scene edge) is measured by the strength of the edges of the IWE.

Finally, an optimization algorithm iterates the above steps until the best parameters are found: 
\begin{equation}
\label{eq:bestParamsOriginal}
\bparams^\ast = \arg\max_{\bparams}\costFunc(\bparams).
\end{equation}
As an intuitive interpretation, when the candidate motion is correct, each warped event overlaps at the reference time $\tref$, resulting in an image that clearly presents the edges at the moment of $\tref$.  
On the other hand, if the candidate motion does not match the actual motion of the object, a low-contrast, blurred IWE is produced.

\subsection{Segmentation using the Calculus of Variations}
\label{sec:method:variation}

\textbf{Accounting for multiple motions in the scene}.
In CMax, the warp $\Warp$ defines the type of motion (motion hypothesis, which is a design choice), that is parametrized by $\bparams$.
For instance, when the entire image plane is assumed to move with a common linear velocity,
the warp is a two-dimensional translation $\bx'_k = \bx_k - (t_k-t_\text{ref})\velflow$
and $\bparams \doteq \velflow = (v_x, v_y)^\top$ is the velocity.
However, when two or more IMOs are present in the scene, a motion hypothesis like the one above, with two degrees of freedom (DOFs), cannot fully explain the event data.
In practice, it is not easy to design a warp that covers a wide range of scenes; this requires warps with many DOFs (e.g., optical flow). 
And it is also not easy to estimate its best parameters $\bparams^\ast$: high-DOF problems are more ill-conditioned than low-DOF ones.
Hence, instead of estimating all motions (DOFs) at once,
we proceed to segment different motions iteratively, 
by determining which parts of the event data fit the simple motion model $\bparams \doteq \velflow$,
and fitting new motion parameters to the residual events that do not align with the previous estimation. 
\begin{algorithm}[t]
\caption{Iterative Motion Segmentation Algorithm}
\label{algo:segmentation}
\begin{algorithmic}[1]
 \renewcommand{\algorithmicrequire}{\textbf{Input:}}
 \renewcommand{\algorithmicensure}{\textbf{Output:}}
 \REQUIRE Events $\cE$, threshold $T$ %
 \ENSURE Motion clusters and best motion parameters $\{\cE_i, \bparams^\ast_i \}_{i=1}^{N_\text{cluster}}$ \\
 \STATE Initialize residual events $\cE^{\text{res}} = \cE$, $N_\text{cluster} = 0$, $i=1$.
 \WHILE {$\cE^{\text{res}} \neq \varnothing$}
 \STATE Estimate motion $\bparams^\ast$ using Contrast Maximization \eqref{eq:bestParamsOriginal} or external sensors (e.g., IMU)
 \STATE Warp events \eqref{eq:warp} $\cE^{\text{res}}$
 \STATE Calculate the gradient \eqref{eq:variation}
 \STATE Obtain the current cluster (segment): $\bparams^\ast_i = \bparams^\ast$ 
 and $ \cE_i = \{e_k\}$ s.t. $\lVert \frac{\partial{\costFunc(\bx_k;\bparams^\ast_i)}}{\partial{\bx_k}} \rVert > T$
 \STATE $N_\text{cluster} \leftarrow N_\text{cluster} + 1$
 \STATE Update events using the residuals $\cE^{\text{res}} \leftarrow \{e_k\}$ s.t. $\lVert \frac{\partial{\costFunc(\bx_k;\bparams^\ast)}}{\partial{\bx_k}} \rVert < T$.
 \STATE $i \leftarrow i+1$
  \ENDWHILE
\end{algorithmic}
\end{algorithm}

\textbf{Proposed Approach}. 
Our method is summarized in \cref{fig:method,algo:segmentation}.
Instead of seeking the optimal motion $\bparams^\ast$,
we assume that the motion parameters $\bparams$ are already known with certain accuracy.
The parameters $\bparams$ can be given either by a step of CMax (see \cref{sec:experim:segmentation} for the results)
or by other sensors (e.g., LiDAR or IMU) (see \cref{sec:experim:detection} for the results).
In either case, we assume $\bparams$ represents the dominant motion for the current event data (e.g., ego-motion--induced events, excluding IMOs).
To define the ``goodness of fit'' of events with respect to the current motion $\bparams$,
we are inspired by the Calculus of Variations \cite{Elsgolc2007}, proposing a variational method to classify events based on the motion parameters. 
In CMax \eqref{eq:bestParamsOriginal}, $\bparams^\ast$ may be found iteratively by following the gradient $\partial\costFunc / \partial\bparams$,
e.g., as in steepest ascent 
$\bparams \leftarrow \bparams + \mu \frac{\partial{\costFunc}}{\partial{\bparams}}$ (with $\mu\geq 0$).

Complementary, the variational approach looks at the sensitivity of the objective function with respect to the events, assuming fixed motion parameters $\bparams$:
\begin{equation}
\label{eq:variation}
\frac{\partial{\costFunc(\bx_k;\bparams)}}{\partial{\bx_k}} 
= \left(\frac{\partial{\costFunc(\bx_k;\bparams)}}{\partial{x_k}}, \frac{\partial{\costFunc(\bx_k;\bparams)}}{\partial{y_k}}\right).
\end{equation}
Intuitively, it evaluates how the coordinates of each event affect the contrast function assuming $\bparams$ is fixed, and it can be leveraged to increase the contrast score. 
If an event is aligned with the point trajectories of the current motion $\bparams$, 
then the variation \eqref{eq:variation} should be large,
while if it is not aligned (e.g., IMOs), the variation should be close to zero.

In practice, to calculate the per-event variation \eqref{eq:variation} 
we rely on automatic differentiation libraries, such as PyTorch.
The magnitude of the variation, $\| \partial{\costFunc(\bx_k;\bparams)}/\partial{\bx_k} \|$, informs about how much the event matches the motion hypothesis $\bparams$.
By thresholding it at some level $T$, we thus classify the events into those agreeing with the current motion and the residuals (\cref{algo:segmentation}, line 6). 
This process of determining the dominant motion and classifying events is subsequently repeated for the residual events (segmentation iteration), 
until the number of residual events becomes negligibly small (i.e., noise).
The threshold $T$ can be predetermined, or decided statistically using Otsu's algorithm \cite{Otsu75tsmc} on the histogram of values $\{ \| \partial{\costFunc}/\partial{\bx_k} \| \}$.
We analyse the histogram and discuss more details in the supplementary.

\emph{Visualization}.
To visualize \eqref{eq:variation}, we may build an image (e.g., akin to the IWE) by taking per-pixel averages, 
e.g., such as the ``mean (magnitude) variation image'' (MVI): 
\begin{equation}
\label{eq:gradientMap}
    \text{MVI}(\cE,\bparams) \doteq \frac1{\numEvents(\bx)} \sum_{k=1}^{\numEvents}
    \left\| \frac{\partial{\costFunc(\bx_k;\bparams)}}{\partial{\bx_k}} \right\| \delta(\bx - \bx'_k),
\end{equation}
where $\numEvents(\bx)\doteq \sum_{k=1} \delta(\bx - \bx'_k)$ is the number of warped events at pixel $\bx$ (the IWE).
An example of the MVI is shown in \cref{fig:method}, colored from yellow (low) to blue (high).

\textbf{Technical comparison with prior work}. 
Similarly to prior work, our method is based on CMax, without requiring external labels (\cref{tab:methodComparison}).
However, there are some differences. 
EMSMC \cite{Stoffregen19iccv} uses event probabilities $p_{kj}$ as weights during IWE calculation, whereas our approach does not; 
our resulting event probabilities are 0 or 1 (the events belong to one cluster or to another).
Another difference with respect to EMSMC is the character of the iterative process. 
Our approach revitalizes the idea of the ``greedy'' strategy of \cite{Mitrokhin18iros}, obtaining the clusters sequentially, in decreasing order of dominance, as measured by the number of events that conform with the newly estimated motion.
This is faster than EMSMC, which follows an EM (i.e., non-greedy) approach: all motion clusters may change in every EM iteration.
Differently from EMSGC \cite{Zhou21tnnls} (also in \cref{tab:methodComparison}), we do not need to build a graph of events and find the best partition of the graph to classify events into the IMOs. The number of clusters is automatically determined via the iteration, instead of by energy minimization.

\section{Experiment}
\label{sec:experiment}

In this section, first 
we describe the datasets (\cref{sec:experim:dataset})
and the assessment metrics as well as hyper-parameters (\cref{sec:experim:evalMetrics}).
Then, we report on two different evaluations:
motion segmentation using self-recorded simple datasets (\cref{sec:experim:segmentation})
and moving-object detection (\cref{sec:experim:detection,sec:experim:additionalData}), both using real-world datasets.

\subsection{Datasets}
\label{sec:experim:dataset}

\textbf{Self-recorded Dataset.} 
To show the efficacy of the proposed method, we first use self-recorded data that consist of simple scenes with different IMO patterns.
The dataset, acquired with a DAVIS346 \cite{Taverni18tcsii}, comprises events and images of $346\times260$ pixels.
The camera is static and looks downward and front-to-parallel to the ground,
capturing multiple objects (such as cylinders and toy trains) moving with approximately constant linear velocities.
We record seven sequences that vary the types of objects,
their motion directions (e.g., parallel, opposite, vertical) and speeds.
The dataset contains over 2 million events in total.

\textbf{EVIMO2} \cite{Burner22evimo2} consists of 41 minutes of data from three event cameras ($640\times480$ pixels), one RGB camera, two IMUs,
and accurate object poses from a motion capture system.
We use three sequences (\emph{scene13\_dyn\_test\_00},
\emph{scene10\_dyn\_train\_00} and \emph{scene15\_dyn\_test\_02})
to evaluate IMOs following previous works \cite{zhou2024jstr}.
Depth and segmentation of the scene, including objects, are provided from the event camera viewpoint at 60 Hz, which we use to warp the initial motion cluster.
For each sequence we select time slices that include one IMO, 
hence we evaluate $1.3$M events over 0.76 seconds.

\textbf{EMSGC} \cite{Zhou21tnnls}
consists of small camera motions and several IMOs in the scene, which are recorded using a hand-held DAVIS346 camera.
We test some real-world sequences following previous work \cite{Zhou21tnnls}.

\textbf{ECD} \cite{Mueggler17ijrr} is a standard dataset for various camera ego-motion estimation \cite{Gallego17ral,Zhu17cvpr,Rosinol18ral,Mueggler18tro}.
Using a DAVIS240C camera ($240 \times 180$ px \cite{Brandli14ssc}),
each sequence provides events, frames, calibration information, IMU data,
and ground truth (GT) camera poses (at $200$~Hz).
We use \emph{dynamic\_rotation} that includes IMOs (a person moving objects).

\subsection{Evaluation Metrics and Hyper-parameters}
\label{sec:experim:evalMetrics}

\textbf{FWL.}
We use the Flow Warp Loss (FWL) \cite{Stoffregen20eccv} for evaluating motion segmentation results.
The FWL is defined as the ratio of the variance of IWE with respect to the variance of the original event image (i.e., IWE of no warp):
\begin{equation}
\mathrm{FWL} \doteq \dfrac{\sigma^2(\IWE(\bx;\bparams))}{\sigma^2(\IWE(\bx;\bzero))}.
\label{eq:FWL}
\end{equation}
It is interpreted as follows:
an $\mathrm{FWL} > 1$ indicates that the output IWE has higher contrast than the input IWE, implying that the events are better aligned and segmented to different motions.
Without the event collapse phenomenon \cite{Shiba22aisy}, 
which is avoided because the warps used are well-behaved \cite{Shiba22sensors}, higher FWL is better.

\textbf{Intersection over Union (IoU)} is a widely used metric for evaluating the accuracy of object detection and segmentation.
It measures the overlap between the predicted segmentation $P$ (motion cluster or its bounding box) and the ground truth $G$.
Specifically, IoU is defined as the ratio of the intersection area of the predicted and ground truth (GT) regions to their union area:
\begin{equation}
\mathrm{IoU} \doteq \dfrac{|P \cap G|}{|P \cup G|}.
\end{equation}
$\text{IoU}=1$ indicates perfect overlap between the prediction and GT, while $\text{IoU}=0$ indicates no overlap.
High IoU values indicate better segmentation or detection accuracy.

\def\figWidth{0.23\linewidth}
\begin{figure}[t]
	\centering
    {\scriptsize
    \setlength{\tabcolsep}{1pt}
	\begin{tabular}{
	>{\centering\arraybackslash}m{0.3cm}
	>{\centering\arraybackslash}m{\figWidth} 
	>{\centering\arraybackslash}m{\figWidth} 
	>{\centering\arraybackslash}m{\figWidth} 
	>{\centering\arraybackslash}m{\figWidth}}
		\\

		\rotatebox{90}{\makecell{cylinder2}}
		&\gframe{\includegraphics[trim={0 0 0 0},clip,width=\linewidth]{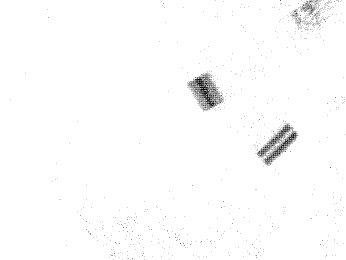}}
		&\gframe{\includegraphics[trim={0 0 0 0},clip,width=\linewidth]{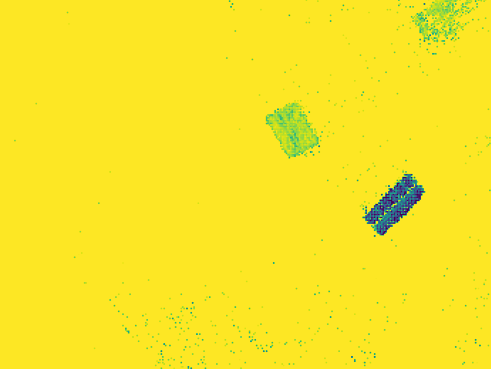}}
		&\gframe{\includegraphics[trim={0 0 0 0},clip,width=\linewidth]{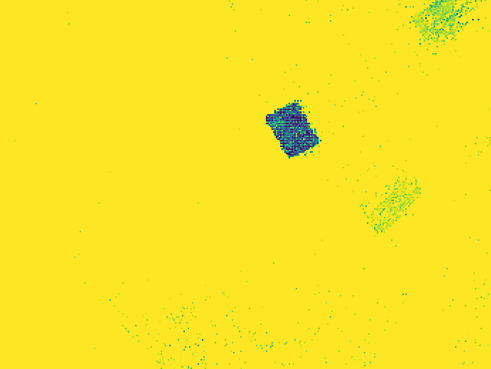}}
		&\gframe{\includegraphics[trim={0 0 0 0},clip,width=\linewidth]{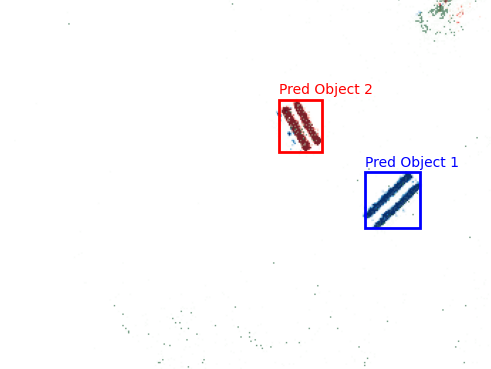}}
		\\

		\rotatebox{90}{\makecell{toy1}}
		&\gframe{\includegraphics[trim={0 0 0 0},clip,width=\linewidth]{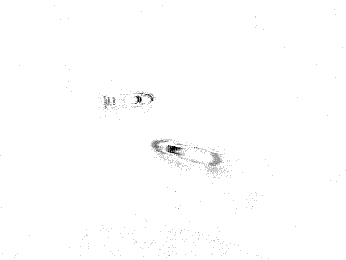}}
		&\gframe{\includegraphics[trim={0 0 0 0},clip,width=\linewidth]{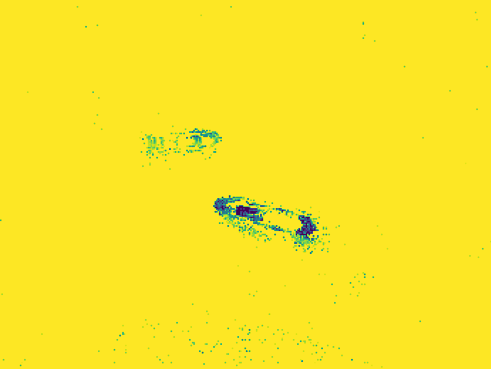}}
		&\gframe{\includegraphics[trim={0 0 0 0},clip,width=\linewidth]{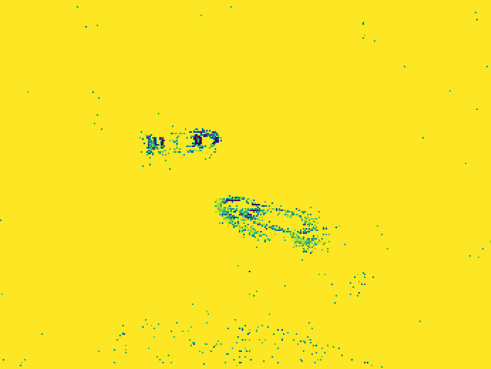}}
		&\gframe{\includegraphics[trim={0 0 0 0},clip,width=\linewidth]{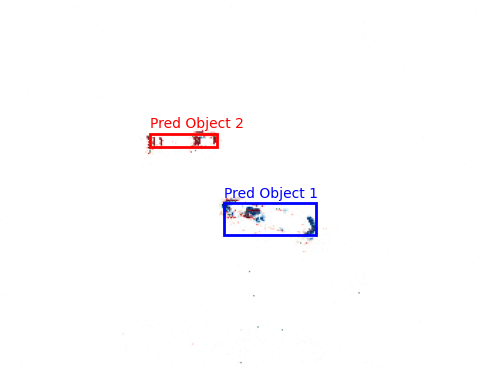}}
		\\

		\rotatebox{90}{\makecell{toy3}}
		&\gframe{\includegraphics[trim={0 0 0 0},clip,width=\linewidth]{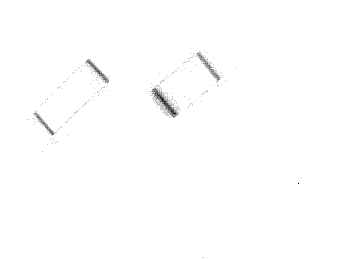}}
		&\gframe{\includegraphics[trim={0 0 0 0},clip,width=\linewidth]{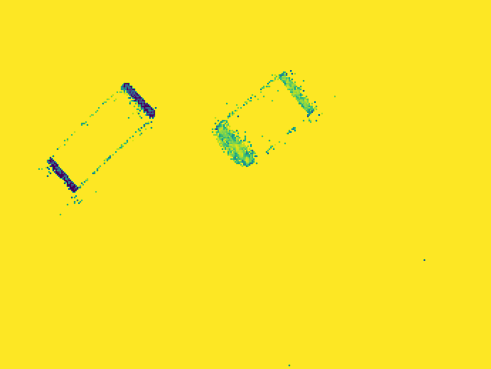}}
		&\gframe{\includegraphics[trim={0 0 0 0},clip,width=\linewidth]{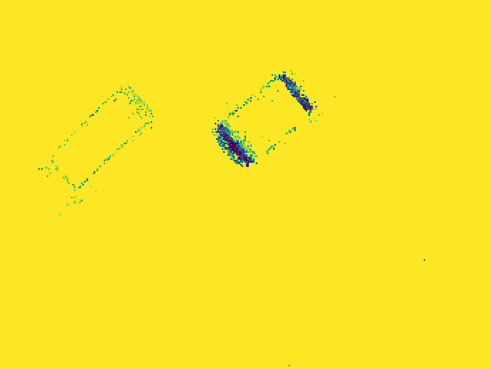}}
		&\gframe{\includegraphics[trim={0 0 0 0},clip,width=\linewidth]{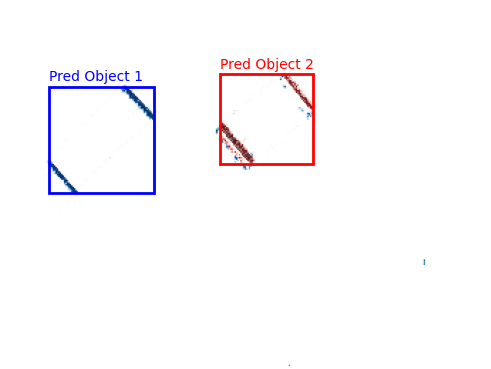}}
		\\

		& (a) Events
        & (b) MVI for the first cluster
		& (c) MVI for the second cluster
  		& (d) Segmentation and Bounding Box
		\\
	\end{tabular}
	}
	\caption{\emph{Iterative clustering on self-recorded data}. 
 (b)-(c): the mean variation images (MVIs) during two iterations show that the variation \eqref{eq:gradientMap} becomes high (i.e., blue) for the remaining IMO (color bar in \cref{fig:method}). 
 We use these heat map values to produce the segmentation results (d).
 }
\label{fig:clustering_ours}
\vspace{-1ex}
\end{figure}

\def\figWidth{0.151\linewidth}
\begin{figure*}[t]
	\centering
    {\scriptsize
    \setlength{\tabcolsep}{1pt}
	\begin{tabular}{
	>{\centering\arraybackslash}m{0.3cm}
	>{\centering\arraybackslash}m{\figWidth} 
	>{\centering\arraybackslash}m{\figWidth} 
	>{\centering\arraybackslash}m{\figWidth} 
	>{\centering\arraybackslash}m{\figWidth} 
	>{\centering\arraybackslash}m{\figWidth}
	>{\centering\arraybackslash}m{\figWidth}}
		\\

		\rotatebox{90}{\makecell{Cylinder1}}
		&\gframe{\includegraphics[trim={0 0 0 0},clip,width=\linewidth]{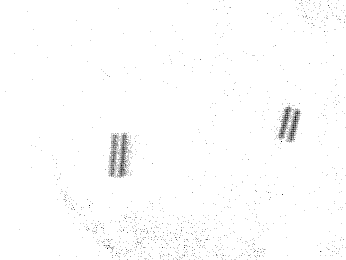}}
		&\gframe{\includegraphics[trim={0 0 0 0},clip,width=\linewidth]{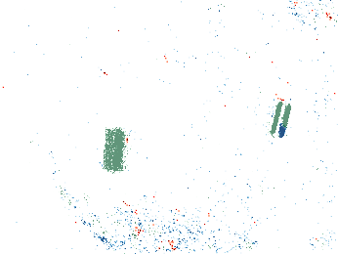}}
            &\gframe{\includegraphics[trim={0 0 0 0},clip,width=\linewidth]{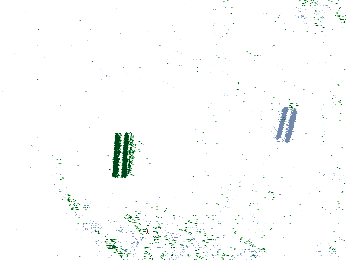}}
            &\gframe{\includegraphics[trim={0 0 0 0},clip,width=\linewidth]{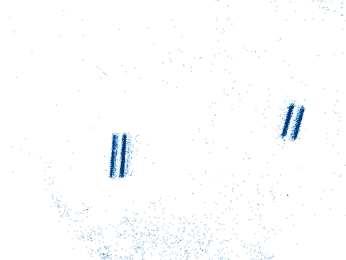}}
		&\gframe{\includegraphics[trim={0 0 0 0},clip,width=\linewidth]{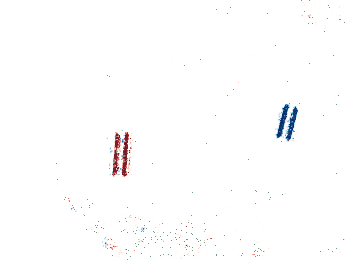}}
          &\gframe{\includegraphics[trim={0 0 0 0},clip,width=\linewidth]{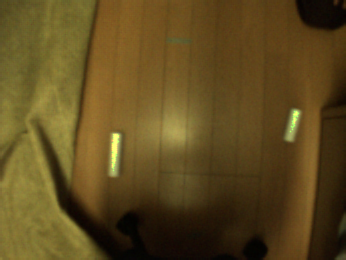}}
            \\

		\rotatebox{90}{\makecell{Cylinder2}}            
            &\gframe{\includegraphics[trim={0 0 0 0},clip,width=\linewidth]{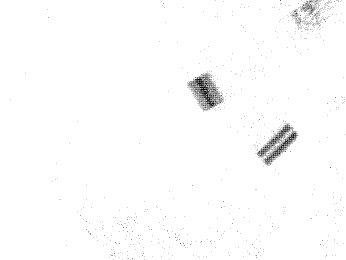}}
		&\gframe{\includegraphics[trim={0 0 0 0},clip,width=\linewidth]{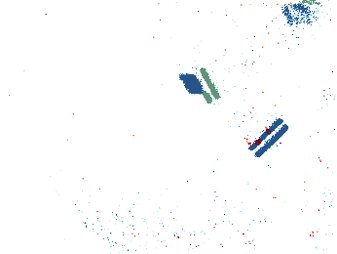}}
            &\gframe{\includegraphics[trim={0 0 0 0},clip,width=\linewidth]{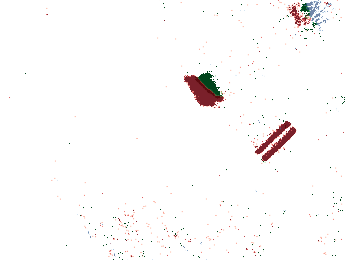}}
            &\gframe{\includegraphics[trim={0 0 0 0},clip,width=\linewidth]{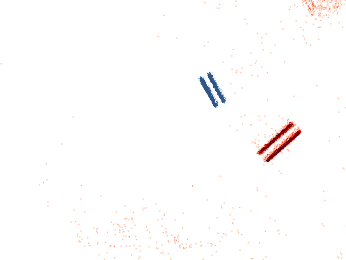}}
		&\gframe{\includegraphics[trim={0 0 0 0},clip,width=\linewidth]{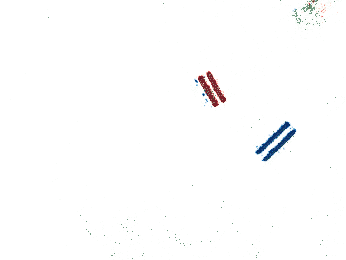}}
          &\gframe{\includegraphics[trim={0 0 0 0},clip,width=\linewidth]{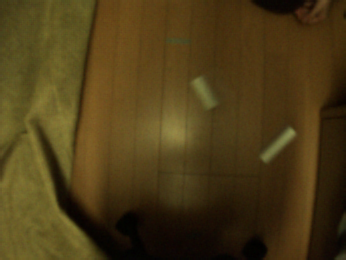}}
            \\

            \rotatebox{90}{\makecell{Toy1}}
            &\gframe{\includegraphics[trim={0 0 0 0},clip,width=\linewidth]{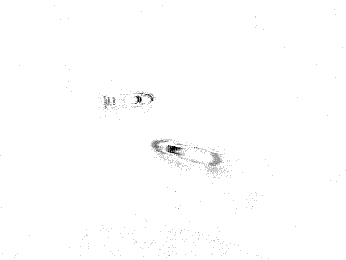}}
		&\gframe{\includegraphics[trim={0 0 0 0},clip,width=\linewidth]{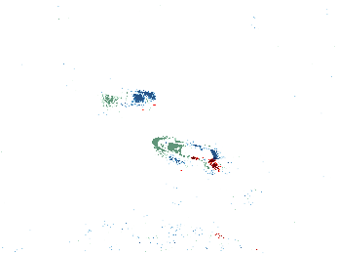}}
		&\gframe{\includegraphics[trim={0 0 0 0},clip,width=\linewidth]{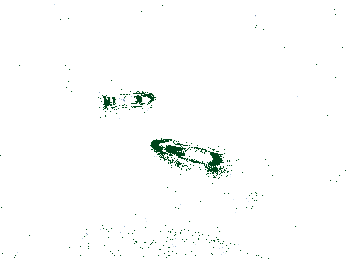}}
            &\gframe{\includegraphics[trim={0 0 0 0},clip,width=\linewidth]{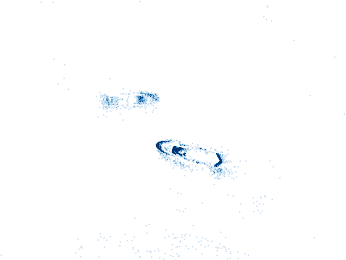}}
		&\gframe{\includegraphics[trim={0 0 0 0},clip,width=\linewidth]{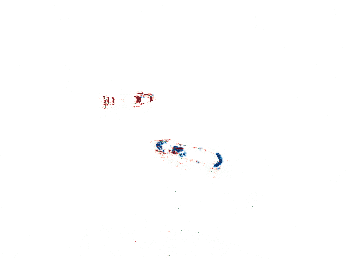}}
          &\gframe{\includegraphics[trim={0 0 0 0},clip,width=\linewidth]{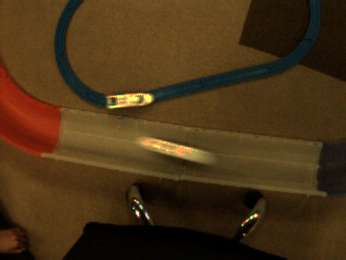}}
            \\

            \rotatebox{90}{\makecell{Toy2}}            
		&\gframe{\includegraphics[trim={0 0 0 0},clip,width=\linewidth]{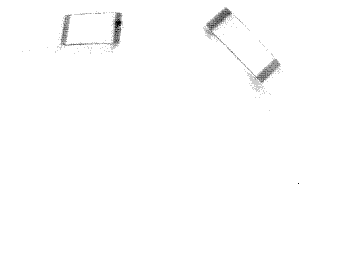}}
		&\gframe{\includegraphics[trim={0 0 0 0},clip,width=\linewidth]{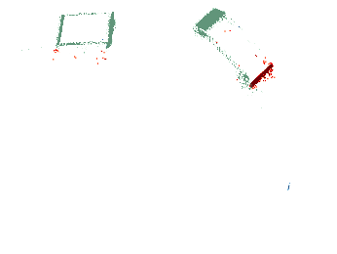}}
		&\gframe{\includegraphics[trim={0 0 0 0},clip,width=\linewidth]{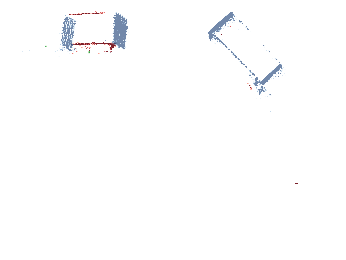}}
            &\gframe{\includegraphics[trim={0 0 0 0},clip,width=\linewidth]{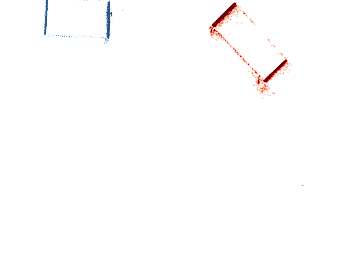}}
		&\gframe{\includegraphics[trim={0 0 0 0},clip,width=\linewidth]{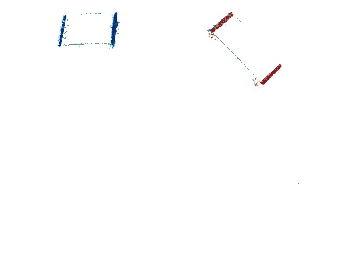}}
        &\gframe{\includegraphics[trim={0 0 0 0},clip,width=\linewidth]{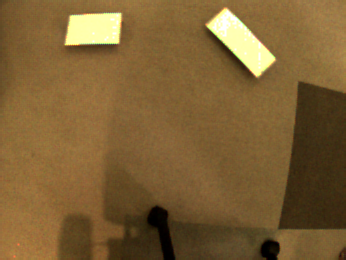}}
            \\

            \rotatebox{90}{\makecell{Toy4}}
		&\gframe{\includegraphics[trim={0 0 0 0},clip,width=\linewidth]{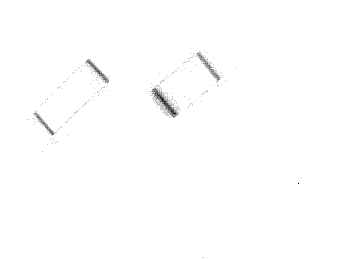}}
		&\gframe{\includegraphics[trim={0 0 0 0},clip,width=\linewidth]{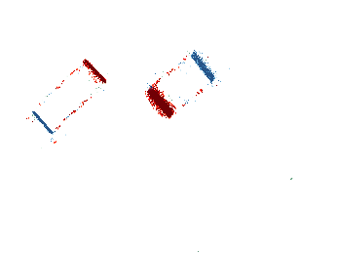}}
  		&\gframe{\includegraphics[trim={0 0 0 0},clip,width=\linewidth]{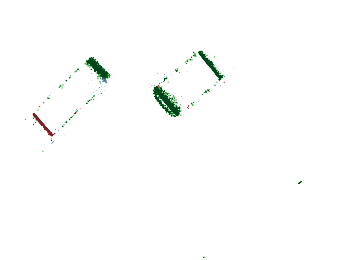}}
            &\gframe{\includegraphics[trim={0 0 0 0},clip,width=\linewidth]{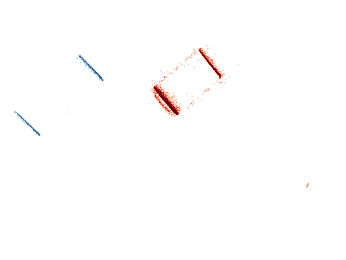}}
		&\gframe{\includegraphics[trim={0 0 0 0},clip,width=\linewidth]{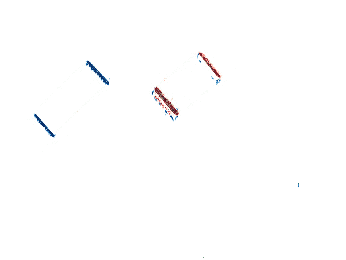}}
          &\gframe{\includegraphics[trim={0 0 0 0},clip,width=\linewidth]{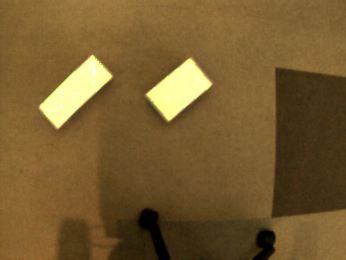}}
            \\

            \rotatebox{90}{\makecell{Toy5}}
		&\gframe{\includegraphics[trim={0 0 0 0},clip,width=\linewidth]{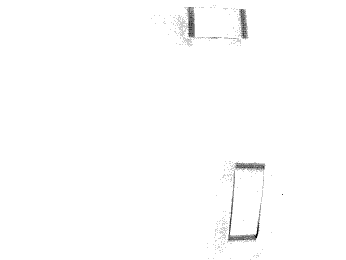}}
		&\gframe{\includegraphics[trim={0 0 0 0},clip,width=\linewidth]{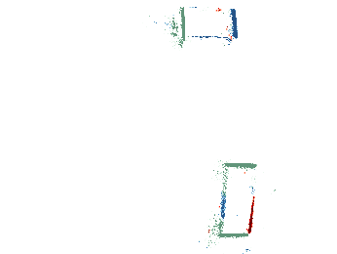}}
		&\gframe{\includegraphics[trim={0 0 0 0},clip,width=\linewidth]{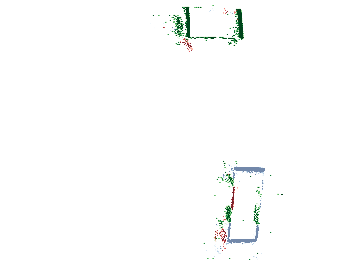}}
            &\gframe{\includegraphics[trim={0 0 0 0},clip,width=\linewidth]{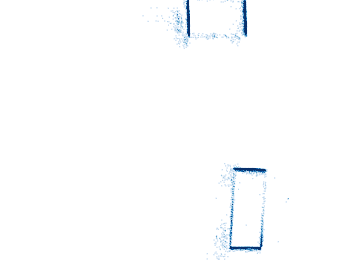}}
		&\gframe{\includegraphics[trim={0 0 0 0},clip,width=\linewidth]{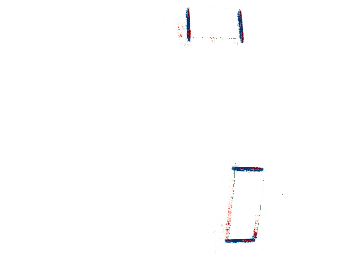}}
      &\gframe{\includegraphics[trim={0 0 0 0},clip,width=\linewidth]{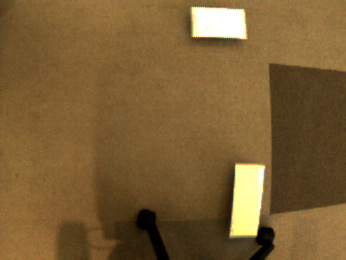}}
            \\

            & (a) Event
		& (b) EMSMC \cite{Stoffregen19iccv} \emph{zero}
		& (c) EMSMC \cite{Stoffregen19iccv} \emph{random}
            & (d) EMSGC \cite{Zhou21tnnls}
		& (e) Ours (\emph{zero})
		& (f) Reference image
		\\
	\end{tabular}
	}
	\caption{\emph{Segmentation results on our dataset}. For benchmark, we compare with \cite{Zhou21tnnls} and \cite{Stoffregen19iccv} using different initialization strategies.
    }
\label{fig:comparison_ours}
\end{figure*}

\begin{table*}[t]
    \centering
    \resizebox{\textwidth}{!}{
    \begin{tabular}{lcccccccccccccc}
        \toprule
         & \multicolumn{2}{c}{cylinder1} & \multicolumn{2}{c}{cylinder2} & \multicolumn{2}{c}{toy1} & \multicolumn{2}{c}{toy2} & \multicolumn{2}{c}{toy3} & \multicolumn{2}{c}{toy4} & \multicolumn{2}{c}{toy5} \\
         \cmidrule(l{1mm}r{1mm}){2-3}
         \cmidrule(l{1mm}r{1mm}){4-5}
         \cmidrule(l{1mm}r{1mm}){6-7}
         \cmidrule(l{1mm}r{1mm}){8-9}
         \cmidrule(l{1mm}r{1mm}){10-11}
         \cmidrule(l{1mm}r{1mm}){12-13}
         \cmidrule(l{1mm}r{1mm}){14-15}

         & FWL & IoU & FWL & IoU & FWL & IoU & FWL & IoU & FWL & IoU & FWL & IoU & FWL & IoU \\
        \midrule
        EMSMC \cite{Stoffregen19iccv} \emph{zero init.} & 1.01 & 0.34 & 1.32 & \unum{1.2}{0.56} & \unum{1.2}{1.12} & 0.18 & 0.94 & 0.62 & 0.94 & 0.28 & 1.31 & 0.62 & \unum{1.2}{1.37} & 0.26 \\
        EMSMC \cite{Stoffregen19iccv} \emph{random init.} & \bnum{1.24} & \unum{1.2}{0.50} & 1.20 & 0.55 & 1.00 & \bnum{0.23} & \bnum{1.96} & \unum{1.2}{0.74} & \unum{1.2}{1.10} & \bnum{0.67} & 1.26 & \unum{1.2}{0.63} & 1.33 & \unum{1.2}{0.27} \\
        EMSGC \cite{Zhou21tnnls} & 1.09 & \novalue & \unum{1.2}{1.33} & \novalue & 1.08 & \novalue & 1.68 & \novalue & 1.06 & \novalue & \unum{1.2}{1.34} & \novalue & 1.18 & \novalue \\
        Ours (\emph{zero init.}) & \unum{1.2}{1.16} & \bnum{0.63} & \bnum{1.40} & \bnum{0.81} & \bnum{1.14} & \unum{1.2}{0.19} & \unum{1.2}{1.79} & \bnum{0.83} & \bnum{1.13} & \unum{1.2}{0.61} & \bnum{1.56} & \bnum{0.73} & \bnum{1.40} & \bnum{0.31} \\
        \bottomrule
    \end{tabular}
    }
    \caption{\emph{Quantitative results on self-recorded dataset}. 
    Higher FWL and IoU values indicate better segmentation. 
    The EM-based algorithm (EMSMC) \cite{Stoffregen19iccv} relies on the initial value of the motion, while ours achieves high FWL and IoU values without initialization (i.e., zero).}
    \label{tab:quatiative_result}
\end{table*}

\textbf{Hyper-parameters}.
We use various motion models for the camera ego-motion to test the efficacy of the proposed method:
$2$-DOF feature flow for the self-recorded and EMSGC datasets where the camera is static (almost static for EMSGC),
use the static-scene depth and ego-motion parameterization (i.e., motion field that has $\numPixels + 6$ DOFs, where $\numPixels$ is the number of pixels) for the EVIMO2 dataset
(see \cref{sec:experim:detection}),
and use the $3$-DOF rotational motion for the ECD dataset.
Also, in one event slice, we use events $10$~ms for the self-recorded dataset, $50$~ms for the EVIMO2 experiments,
and Adam \cite{Kingma15iclr} with the learning rate of $0.5$ as the optimizer.
We leave the further parameter details to the released implementation.

\def\figWidth{0.159\linewidth}
\begin{figure*}[t!]
	\centering
    {\scriptsize
    \setlength{\tabcolsep}{1pt}
	\begin{tabular}{
	>{\centering\arraybackslash}m{0.3cm}
	>{\centering\arraybackslash}m{\figWidth} 
	>{\centering\arraybackslash}m{\figWidth} 
	>{\centering\arraybackslash}m{\figWidth} 
	>{\centering\arraybackslash}m{\figWidth}
	>{\centering\arraybackslash}m{\figWidth}
	>{\centering\arraybackslash}m{\figWidth}}
		\\

		\rotatebox{90}{\makecell{scene13\_dyn\_test\_00}}
		&\gframe{\includegraphics[trim={0 0 0 0},clip,width=\linewidth]{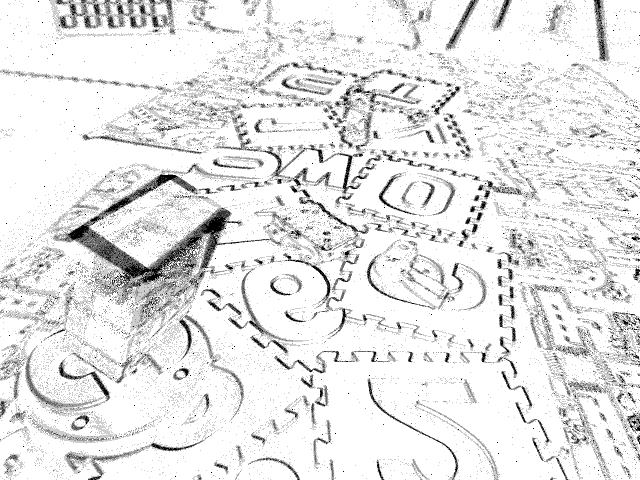}}
		&\gframe{\includegraphics[trim={0 0 0 0},clip,width=\linewidth]{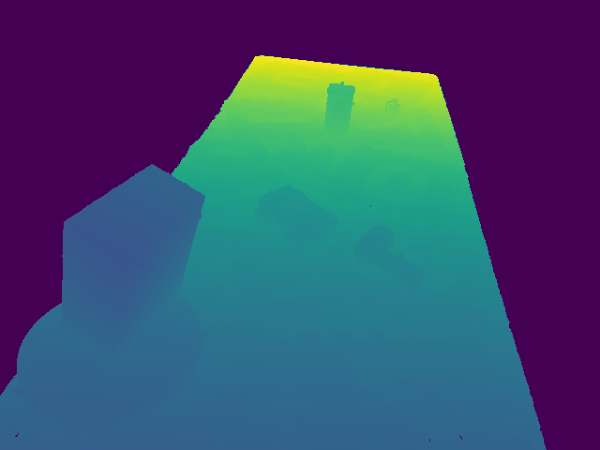}}
		&\gframe{\includegraphics[trim={0 0 0 0},clip,width=\linewidth]{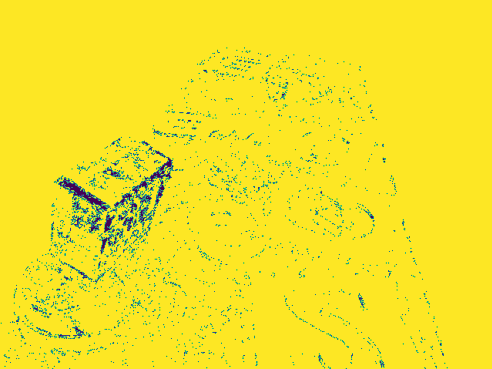}}
		&\gframe{\includegraphics[trim={0 0 0 0},clip,width=\linewidth]{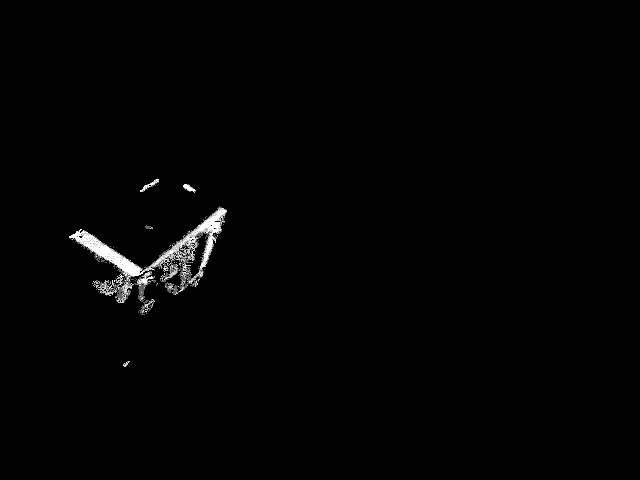}}
		&\gframe{\includegraphics[trim={0 0 0 0},clip,width=\linewidth]{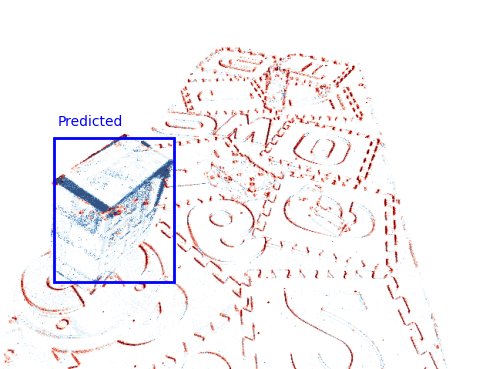}}
		&\gframe{\includegraphics[trim={0 0 0 0},clip,width=\linewidth]{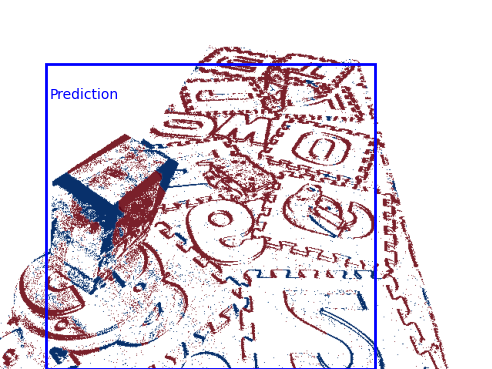}}
		\\

		\rotatebox{90}{\makecell{scene15\_dyn\_test\_02}}
		&\gframe{\includegraphics[trim={0 0 0 0},clip,width=\linewidth]{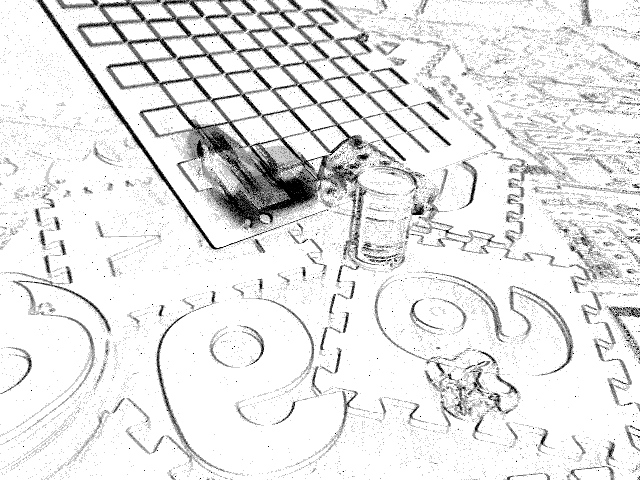}}
		&\gframe{\includegraphics[trim={0 0 0 0},clip,width=\linewidth]{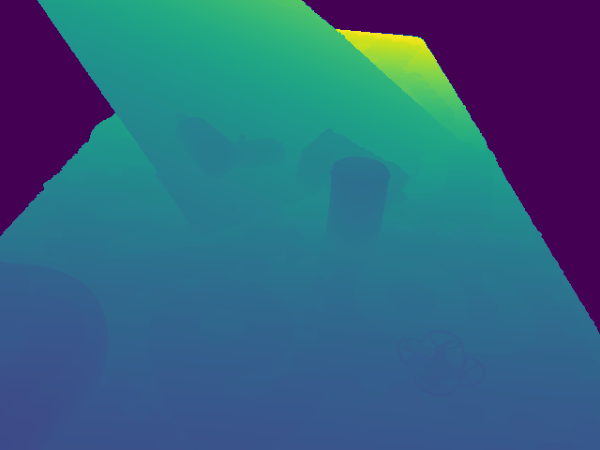}}
		&\gframe{\includegraphics[trim={0 0 0 0},clip,width=\linewidth]{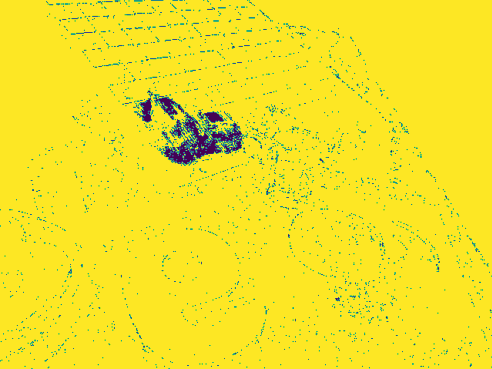}}
		&\gframe{\includegraphics[trim={0 0 0 0},clip,width=\linewidth]{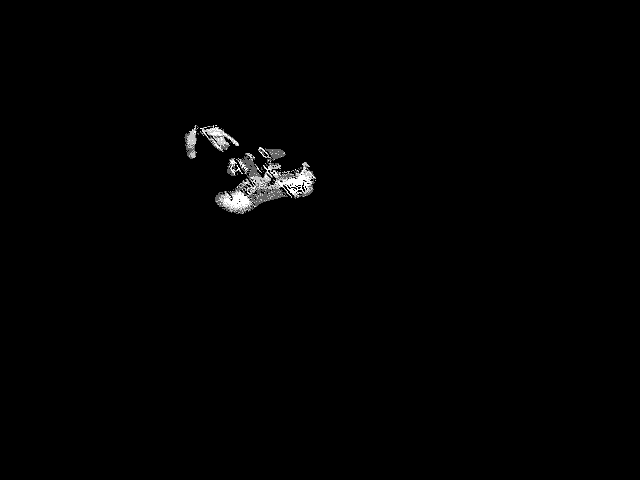}}
		&\gframe{\includegraphics[trim={0 0 0 0},clip,width=\linewidth]{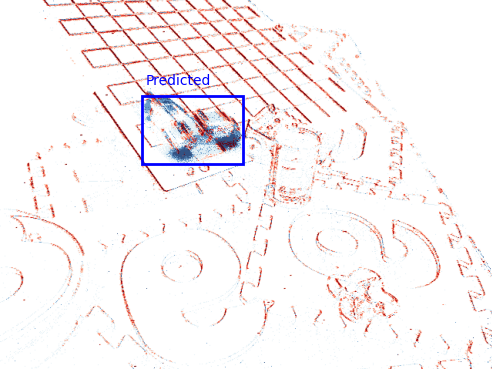}}
		&\gframe{\includegraphics[trim={0 0 0 0},clip,width=\linewidth]{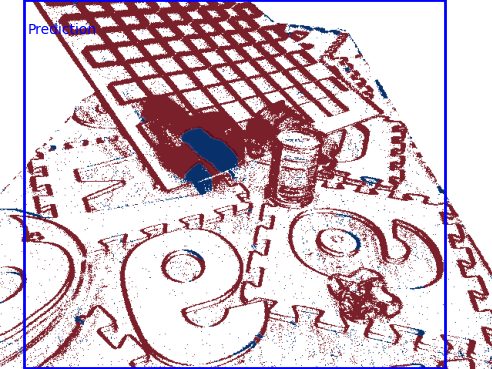}}
		\\

		\rotatebox{90}{\makecell{scene10\_dyn\_train\_00}}
		&\gframe{\includegraphics[trim={0 0 0 0},clip,width=\linewidth]{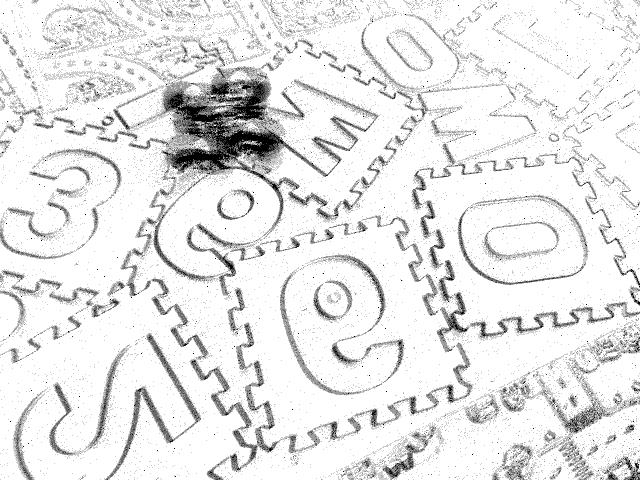}}
		&\gframe{\includegraphics[trim={0 0 0 0},clip,width=\linewidth]{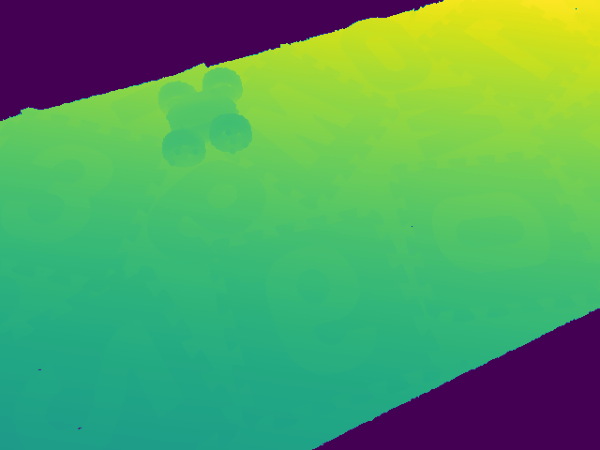}}
		&\gframe{\includegraphics[trim={0 0 0 0},clip,width=\linewidth]{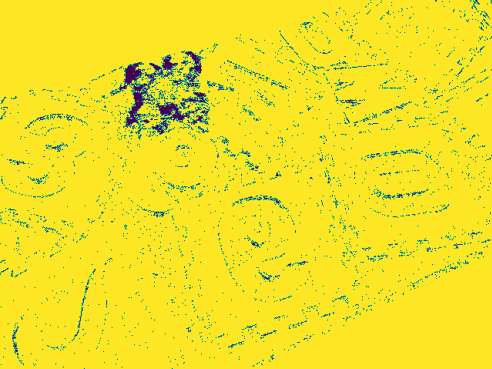}}
		&\gframe{\includegraphics[trim={0 0 0 0},clip,width=\linewidth]{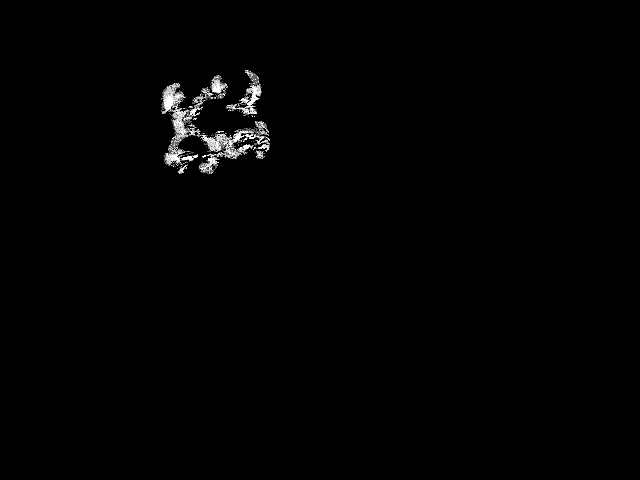}}
		&\gframe{\includegraphics[trim={0 0 0 0},clip,width=\linewidth]{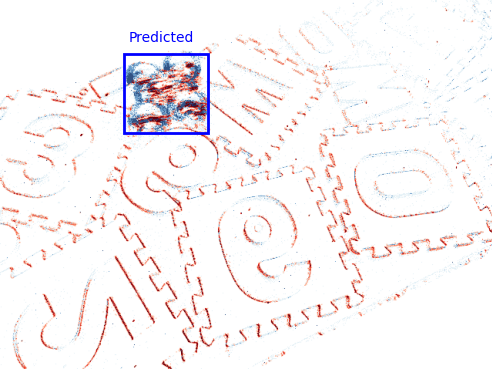}}
		&\gframe{\includegraphics[trim={0 0 0 0},clip,width=\linewidth]{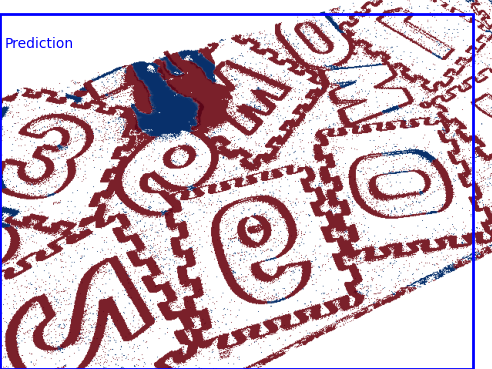}}
		\\

		& (a) Event
        & (b) GT Depth
		& (c) MVI
		& (d) Object Mask
  		& (e) Segmentation and Bounding Box
        & (f) Reference results by EMSMC \cite{Stoffregen19iccv}
		\\
	\end{tabular}
	}
	\caption{\emph{Motion segmentation and moving-object detection results on the EVIMO2 dataset}. Instead of estimating the motion parameter $\bparams$, first we use the GT depth and IMU to obtain the warp for the static parts of the scene. 
    As the MVI (c) clearly shows, the ``residual'' agree with the IMOs. 
    Our segmentation results successfully detect the IMOs (e), compared with the baseline method \cite{Stoffregen19iccv}.
    }
\label{fig:detection_evimo}
\vspace{-1ex}
\end{figure*}

\subsection{Motion Segmentation}
\label{sec:experim:segmentation}

\Cref{fig:clustering_ours} shows the results of the iterative motion segmentation on the self-recorded dataset.
The mean variation images (MVIs) clearly show that the first variation becomes small when the events are aligned with the motion parameter (b, c). %
The agreement between the obtained segmentation results and the bounding boxes of the moving objects demonstrates the capabilities of our method to accurately classify IMOs (d).

The comparison with previous works on motion segmentation is shown in \cref{fig:comparison_ours,tab:quatiative_result}.
The EMSMC method \cite{Stoffregen19iccv} is highly dependent on the initial motion parameters for each cluster, resulting in unstable outcomes across different sequences. 
In the experiments,  we test two types of initialization strategies of the motion parameters: 
zero initialization and random initialization, while our method is tested with zero initialization.
We also compare with EMSGC.

We find that the initialization at zero is challenging for EMSMC \cite{Stoffregen19iccv}.
However, regardless of the initialization strategy, 
the proposed method consistently provides favorable FWL scores and segmentation results, showcasing the effectiveness of the iterative segmentation approach.

\def\figWidth{0.44\linewidth}
\begin{figure}[t!]
	\centering
    {\scriptsize
    \setlength{\tabcolsep}{1pt}
	\begin{tabular}{
	>{\centering\arraybackslash}m{\figWidth} 
	>{\centering\arraybackslash}m{\figWidth}
	>{\centering\arraybackslash}m{\figWidth}}
		\\

		\gframe{\includegraphics[trim={0 0 0 0},clip,width=\linewidth]{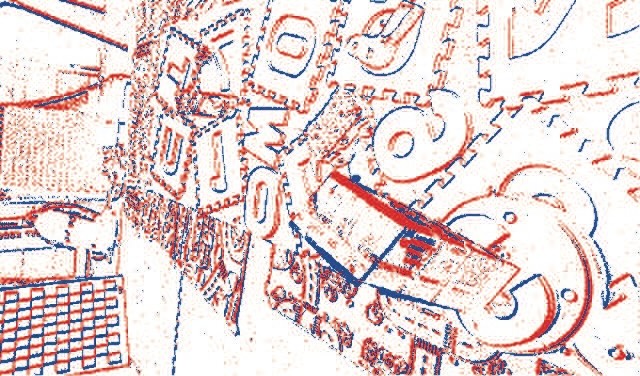}}
	&\gframe{\includegraphics[trim={0 0 0 0},clip,width=\linewidth]{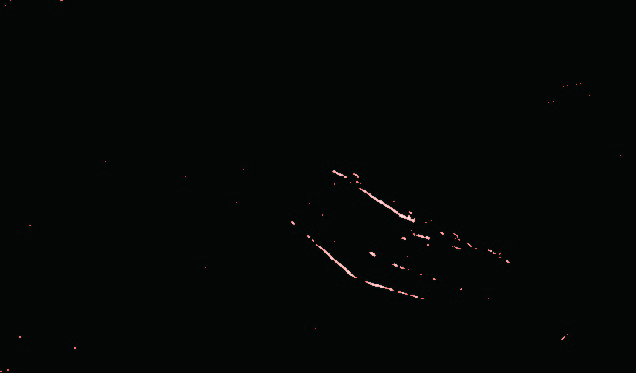}}
        \\        
        (a) Events
	& (b) Mitrokhin et al. \cite{Mitrokhin18iros}
        \\
        \gframe{\includegraphics[trim={0 0 0 0},clip,width=\linewidth]{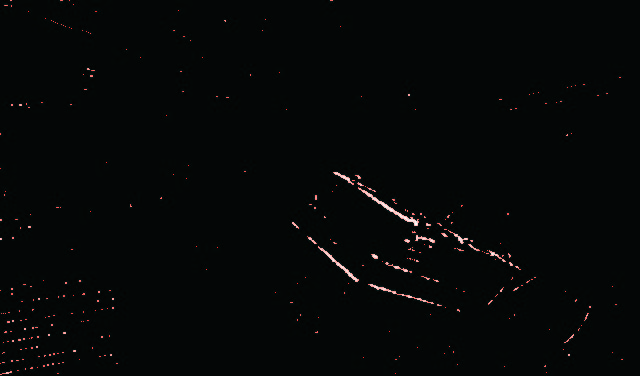}}
	&\gframe{\includegraphics[trim={0 0 0 0},clip,width=\linewidth]{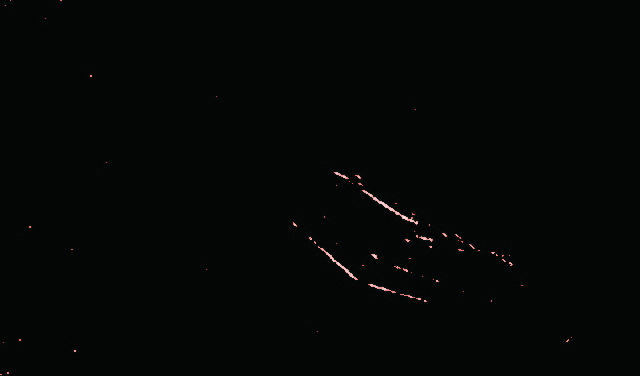}}
        \\
        (c) EMSCG \cite{Stoffregen19iccv}
	& (d) Zhao et al. \cite{zhao23icra}
        \\
        \gframe{\includegraphics[trim={0 0 0 0},clip,width=\linewidth]{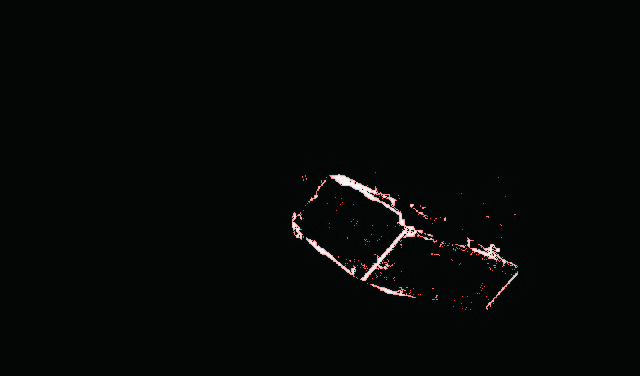}}
        & \gframe{\includegraphics[trim={0 0 0 0},clip,width=\linewidth]{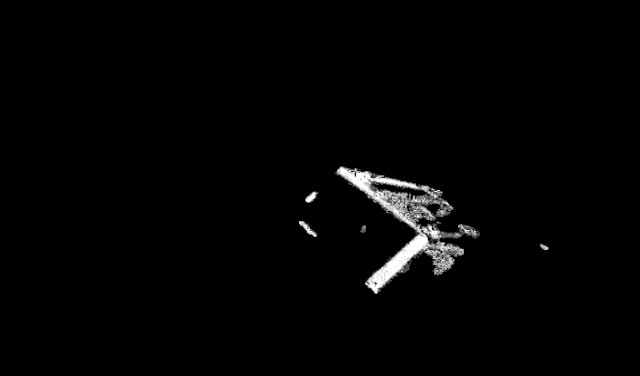}}
        \\
	(e) Zhou et al. \cite{zhou2024jstr}
	& (f) Ours
	\\
	\end{tabular}
	}
	\caption{\emph{Comparison with prior work on the EVIMO2 dataset}. Non-black pixels in each method are IMO.
    The figure is generated mimicking that in \cite{zhou2024jstr}.
    }
\label{fig:comparison_evimo}
\end{figure}

\sisetup{round-mode=places,round-precision=2}
\begin{table}[t!]
\centering
\adjustbox{max width=.6\columnwidth}{%
\setlength{\tabcolsep}{4pt}
\begin{tabular}{l*{3}{S}}
\toprule
Method & \text{IoU($\uparrow$)} & \text{FWL($\uparrow$)} \\  
\midrule
Mitrokhin \cite{Mitrokhin18iros} & 0.61 & \novalue \\
EMSGC \cite{Zhou21tnnls} & 0.63 & \novalue \\
Zhao \cite{zhao23icra} & 0.62 & \novalue \\
Zhou \cite{zhou2024jstr} & 0.64 & \novalue \\
EMSMC \cite{Stoffregen19iccv} & 0.16 & 1.63\\
Ours & \bnum{0.84} & \bnum{2.25} \\ 
\bottomrule
\end{tabular}
}
\caption{\label{tab:resultsevimo}\emph{Quantitative results of moving-object detection on EVIMO2 datasets}. The proposed method produces the highest IoU among the methods compared.}
\vspace{-2ex}
\end{table}

\subsection{Moving Object Detection}
\label{sec:experim:detection}

\Cref{fig:detection_evimo} shows moving object detection results on the EVIMO2 dataset \cite{Burner22evimo2}.
Here, we assume that the calibration parameters are known.
Using external inputs (IMU and GT depth),
we estimate the motion of the background events and compute the MVI of the scene, which generates clear object masks for the IMOs.
Following the well-known motion field equation \cite{Trucco98book},
the motion of the static background can be estimated
using the scene depth $\depth(\bx)$ and the camera motion (linear velocity $\linvel$ and angular velocity $\angvel$),
$\velflow(\bx) = \frac1{\depth(\bx)}A(\bx)\linvel + B(\bx)\angvel$.
Alternatively, one can estimate the depth and camera motion via CMax (e.g., \cite{Shiba24pami}), which we leave for future work.
Since the background events are aligned through motion estimation, 
the IMOs have lower first variation, which we use to make object masks.
The experimental results demonstrate that
our method produces accurate segmentation and bounding box results in the three scenes.
While the benchmark method struggles with segmentation in complex scenes with a large number of events,
the proposed method exhibits significantly better detection of moving objects within the scenes.

\Cref{fig:comparison_evimo} and \cref{tab:resultsevimo} show the comparison of moving-object detection versus prior works.
The IoU is calculated using the average value across scenes containing moving objects.
The proposed method achieves state-of-the-art (high) IoU values,
improving by over $30\%$.
This is also confirmed by the FWL values.

\subsection{Result on the Real-world Datasets}
\label{sec:experim:additionalData}

\Cref{fig:qualitative_timodata} provides further motion segmentation results on real-world sequences, from EMSGC \cite{Zhou21tnnls} and ECD \cite{Mueggler17ijrr}. 
The EMSGC plots (first row) show the classification of the events into three clusters:
background induced by the translational motion of the camera, and two pedestrians moving in opposite directions (along with their corresponding bounding boxes).
The ECD plots (second row) show the classification of the events into two clusters: the rotational motion of the camera and the dynamic motion of a person standing up.
The results demonstrate that each object is successfully detected, 
showcasing that our method performs well on real-world data,
thus indicating the potential for various applications.

However, segmentation accuracy deteriorates due to the presence of a considerable amount of event noise, resulting in small bounding boxes for IMOs in the examples.
We do not fine-tune the hyperparameters of the algorithm to improve the results for these specific scenes and leave additional robustness against noise for future work.

\section{Limitations}
\label{sec:limitations}

One of the primary limitations of the proposed method is its processing time.
The method segments multiple motion clusters through an iterative process,
leading to an increase in processing time proportional to the number of iterations (i.e., the number of distinct motions).
Since the original CMax has a complexity of $O(\numEvents + \numPixels)$,
the proposed method does not run in real-time.
The processing times of different methods for the cylinder example ($3$ clusters, $4$k events) on a standard CPU are: $10.27$~s with the proposed method (python), $4.15$~s with EMSMC (python), and $1.88$~s with EMSGC (C++).

Also, %
classifying motions whose direction is similar is challenging.
If the scene has two similar motions, 
the MVI \eqref{eq:variation} becomes similar among these events.
Consequently, 
it becomes challenging to accurately classify their motions.
For example, in \cref{fig:comparison_ours} (Toy 5), two objects exhibit similar motions, making it difficult to segment them.
In principle, this is a matter of the task definition (they cannot be separated just by motion, and additional cues, such as pixel distance, need to be used to segment them).

\def\figWidth{0.49\linewidth}
 \begin{figure}[t]
	\centering
    {\scriptsize
    \setlength{\tabcolsep}{1pt}
	\begin{tabular}{
	>{\centering\arraybackslash}m{\figWidth} 
	>{\centering\arraybackslash}m{\figWidth}
	}  
	    \gframe{\includegraphics[width=\linewidth]{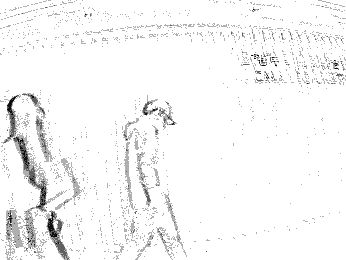}}
	    &\gframe{\includegraphics[width=\linewidth]{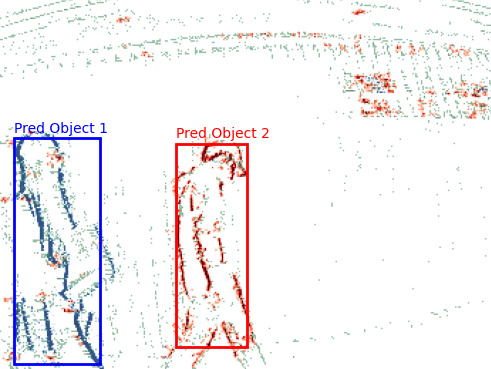}}
		\\

	    \gframe{\includegraphics[width=\linewidth]{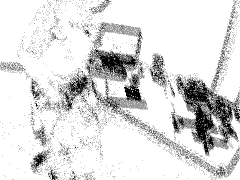}}
	    &\gframe{\includegraphics[width=\linewidth]{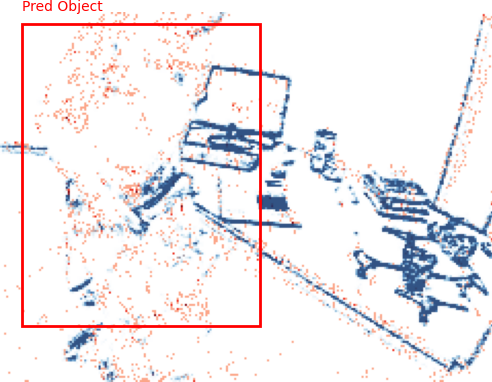}}
		\\

		(a) Input Events
		& (b) Segmentation and BBox
		\\
	\end{tabular}
	}
	\caption{\emph{Results on two real-world datasets} using EMSGC \cite{Zhou21tnnls} and ECD \cite{Mueggler17ijrr}.
    Due to noise, the bounding boxes become small.
	}
\label{fig:qualitative_timodata}
\end{figure}

\section{Conclusion}

In this paper,
we proposed a novel iterative motion segmentation method,
based on the calculus of variations extending the Contrast Maximization framework.
By classifying events into a given motion hypothesis and its residuals,
the proposed method enables motion segmentation and moving-object detection without relying on initialization.
Experimental results show that the proposed method successfully classifies event clusters both for simple datasets and publicly available datasets,
producing sharp, motion-compensated images of warped events.
Also, it achieves state-of-the-art accuracy for moving object detection benchmarks, with an improvement of over $30\%$.
We hope this work broadens the capability of the CMax framework for motion segmentation,
thus contributing to the theoretical advancements in event-based motion estimation.

\ifarxiv

\section*{Acknowledgments}
This research was funded by
the Deutsche Forschungsgemeinschaft (DFG, German Research Foundation) under Germany’s Excellence Strategy – EXC 2002/1 ``Science of Intelligence'' – project number 390523135.

\section{Supplementary}
\label{sec:suppl}

\subsection{Video}
We encourage readers to watch the accompanying video,
which explains the method more intuitively than the still images in the paper.

\subsection{Analysis of Variations}

\Cref{fig:variationsHist} shows an example histogram of variation magnitudes per event,
using the box scene from EVIMO2 
(\emph{scene13\_dyn\_test\_00} in \cref{fig:detection_evimo}).
The order of raw variations is $10^{-5}$, 
while we scale them to $[0, 255]$ for visualization as an MVI.

\def\figWidth{\linewidth}
\begin{figure}[h!]
    \centering
    \includegraphics[width=\figWidth]{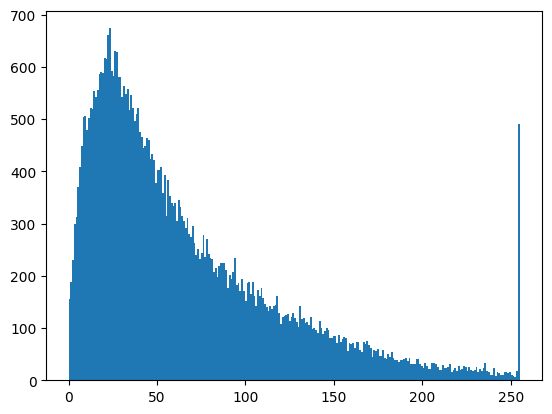}
    \caption{Histogram of the magnitude of variations per event in one sequence example from EVIMO2.
    The magnitude is scaled to $[0, 255]$ for visualization as an image (i.e., MVI).}
\label{fig:variationsHist}
\end{figure}

\newpage
\section{Sensitivity Analysis}
\label{sec:ablation}

In the datasets collected from real-world environments (\cref{sec:experim:detection,sec:experim:additionalData}),
significantly more noise is present compared to ideal conditions.
The noise yields remnants of first variations being introduced during the separation of estimated motion,
and hence, blurs the boundaries of motion segmentation.
To address this, one may apply a simple denoising filter using a Gaussian kernel on the IWE and classify using a certain threshold.

We evaluate the effect of the Gaussian intensity \(\sigma\) and the threshold of the filter on the bounding box accuracy in \cref{fig:thr_sigma}, using the EVIMO2 dataset.
The experimental results show that the IoU consistently remains above 0.7 for the tested parameters (see also \cref{sec:experim:detection}).
We confirm that the proposed method consistently achieves the detection accuracy, even with inversely adjusting $\sigma$ and the threshold.

\def\figWidth{\linewidth}
\begin{figure}[ht]
    \centering
    \includegraphics[width=\figWidth]{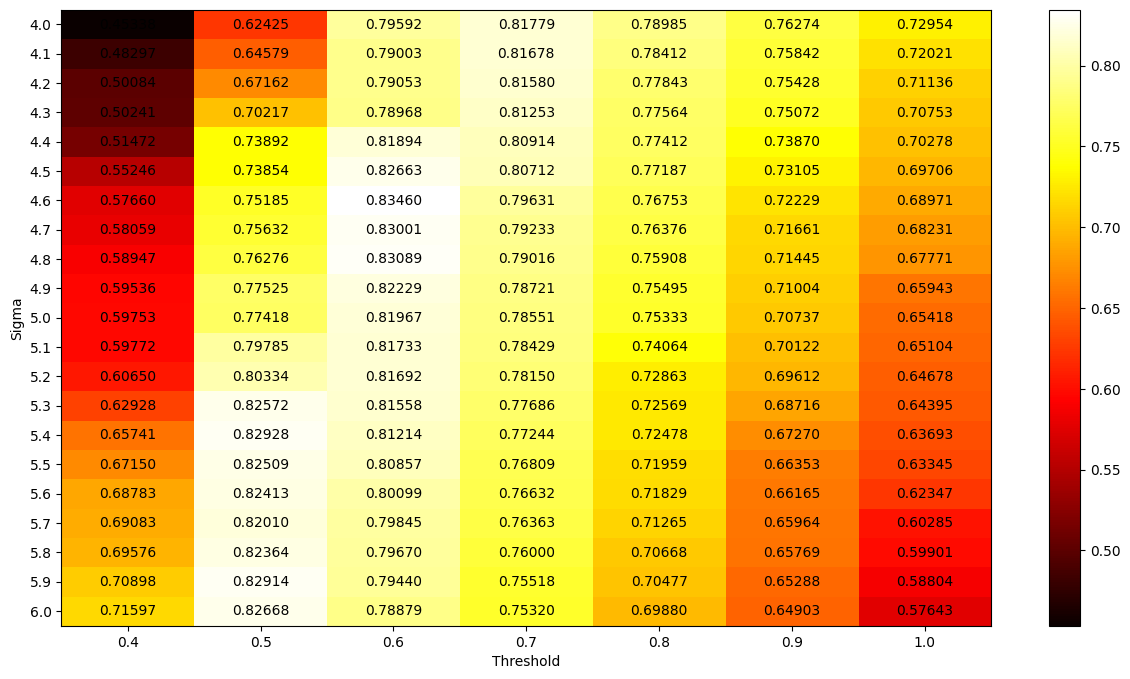}
    \caption{Sensitivity study on the effect of threshold and sigma.}
\label{fig:thr_sigma}
\end{figure}

\clearpage
{
    \small
    \bibliographystyle{ieeenat_fullname}
    \bibliography{all}
}

\else 

{
    \small
    \bibliographystyle{ieeenat_fullname}
    \bibliography{all}

\begin{thebibliography}{48}
\providecommand{\natexlab}[1]{#1}
\providecommand{\url}[1]{\texttt{#1}}
\expandafter\ifx\csname urlstyle\endcsname\relax
  \providecommand{\doi}[1]{doi: #1}\else
  \providecommand{\doi}{doi: \begingroup \urlstyle{rm}\Url}\fi

\bibitem[Alkendi et~al.(2024)Alkendi, Azzam, Javed, Seneviratne, and
  Zweiri]{Alkendi24tim}
Yusra Alkendi, Rana Azzam, Sajid Javed, Lakmal Seneviratne, and Yahya Zweiri.
\newblock Neuromorphic vision-based motion segmentation with graph transformer
  neural network.
\newblock 2024.

\bibitem[Alkendi et~al.(2025)Alkendi, Azzam, Javed, Seneviratne, and
  Zweiri]{Alkendi25tmm}
Yusra Alkendi, Rana Azzam, Sajid Javed, Lakmal Seneviratne, and Yahya Zweiri.
\newblock Neuromorphic vision-based motion segmentation with graph transformer
  neural network.
\newblock \emph{{IEEE} Trans. Multimedia}, 27:\penalty0 385--400, 2025.

\bibitem[Arja et~al.(2024)Arja, Marcireau, Afshar, Ramesh, and
  Cohen]{Arja24arxiv}
Sami Arja, Alexandre Marcireau, Saeed Afshar, Bharath Ramesh, and Gregory
  Cohen.
\newblock Motion segmentation for neuromorphic aerial surveillance.
\newblock \emph{ar{X}iv e-prints}, 2024.

\bibitem[Brandli et~al.(2014)Brandli, Berner, Yang, Liu, and
  Delbruck]{Brandli14ssc}
Christian Brandli, Raphael Berner, Minhao Yang, Shih-Chii Liu, and Tobi
  Delbruck.
\newblock A 240x180 130{dB} 3$\mu$s latency global shutter spatiotemporal
  vision sensor.
\newblock \emph{{IEEE} J. Solid-State Circuits}, 49\penalty0 (10):\penalty0
  2333--2341, 2014.

\bibitem[Burner et~al.(2022)Burner, Mitrokhin, Ferm\"uller, and
  Aloimonos]{Burner22evimo2}
Levi Burner, Anton Mitrokhin, Cornelia Ferm\"uller, and Yiannis Aloimonos.
\newblock {EVIMO2}: An event camera dataset for motion segmentation, optical
  flow, structure from motion, and visual inertial odometry in indoor scenes
  with monocular or stereo algorithms.
\newblock \emph{ar{X}iv e-prints}, 2022.

\bibitem[Elsgolc(2007)]{Elsgolc2007}
Lev~D Elsgolc.
\newblock \emph{Calculus of Variations}.
\newblock Dover Publications, Mineola, NY, 2007.

\bibitem[Gallego and Scaramuzza(2017)]{Gallego17ral}
Guillermo Gallego and Davide Scaramuzza.
\newblock Accurate angular velocity estimation with an event camera.
\newblock \emph{{IEEE} Robot. Autom. Lett.}, 2\penalty0 (2):\penalty0 632--639,
  2017.

\bibitem[Gallego et~al.(2018)Gallego, Rebecq, and Scaramuzza]{Gallego18cvpr}
Guillermo Gallego, Henri Rebecq, and Davide Scaramuzza.
\newblock A unifying contrast maximization framework for event cameras, with
  applications to motion, depth, and optical flow estimation.
\newblock In \emph{{IEEE} Conf. Comput. Vis. Pattern Recog. (CVPR)}, pages
  3867--3876, 2018.

\bibitem[Gallego et~al.(2019)Gallego, Gehrig, and Scaramuzza]{Gallego19cvpr}
Guillermo Gallego, Mathias Gehrig, and Davide Scaramuzza.
\newblock Focus is all you need: Loss functions for event-based vision.
\newblock In \emph{{IEEE} Conf. Comput. Vis. Pattern Recog. (CVPR)}, pages
  12272--12281, 2019.

\bibitem[Gallego et~al.(2022)Gallego, Delbruck, Orchard, Bartolozzi, Taba,
  Censi, Leutenegger, Davison, Conradt, Daniilidis, and
  Scaramuzza]{Gallego20pami}
Guillermo Gallego, Tobi Delbruck, Garrick Orchard, Chiara Bartolozzi, Brian
  Taba, Andrea Censi, Stefan Leutenegger, Andrew Davison, J{\"o}rg Conradt,
  Kostas Daniilidis, and Davide Scaramuzza.
\newblock Event-based vision: A survey.
\newblock \emph{{IEEE} Trans. Pattern Anal. Mach. Intell.}, 44\penalty0
  (1):\penalty0 154--180, 2022.

\bibitem[Georgoulis et~al.(2024)Georgoulis, Ren, Bochicchio, Eckert, Li, and
  Gawel]{Georgoulis24threedv}
Stamatios Georgoulis, Weining Ren, Alfredo Bochicchio, Daniel Eckert, Yuanyou
  Li, and Abel Gawel.
\newblock Out of the room: Generalizing event-based dynamic motion segmentation
  for complex scenes.
\newblock In \emph{Int. Conf. 3D Vision (3DV)}, pages 442--452, 2024.

\bibitem[Gu et~al.(2021)Gu, Learned-Miller, Sheldon, Gallego, and
  Bideau]{Gu21iccv}
Cheng Gu, Erik Learned-Miller, Daniel Sheldon, Guillermo Gallego, and Pia
  Bideau.
\newblock The spatio-temporal {P}oisson point process: A simple model for the
  alignment of event camera data.
\newblock In \emph{Int. Conf. Comput. Vis. (ICCV)}, pages 13495--13504, 2021.

\bibitem[Guo and Gallego(2024)]{Guo24tro}
Shuang Guo and Guillermo Gallego.
\newblock {CMax}-{SLAM}: Event-based rotational-motion bundle adjustment and
  {SLAM} system using contrast maximization.
\newblock \emph{{IEEE} Trans. Robot.}, pages 1--20, 2024.

\bibitem[Guo et~al.(2025)Guo, Hamann, and Gallego]{Guo25e2fai}
Shuang Guo, Friedhelm Hamann, and Guillermo Gallego.
\newblock Unsupervised joint learning of optical flow and intensity with event
  cameras.
\newblock \emph{ar{X}iv e-prints}, 2025.

\bibitem[Hamann et~al.(2024)Hamann, Wang, Asmanis, Chaney, Gallego, and
  Daniilidis]{Hamann24eccv}
Friedhelm Hamann, Ziyun Wang, Ioannis Asmanis, Kenneth Chaney, Guillermo
  Gallego, and Kostas Daniilidis.
\newblock Motion-prior contrast maximization for dense continuous-time motion
  estimation.
\newblock In \emph{Eur. Conf. Comput. Vis. (ECCV)}, pages 18--37, 2024.

\bibitem[Jiang et~al.(2024)Jiang, Moreau, and Davoine]{Jiang24visapp}
Chenao Jiang, Julien Moreau, and Franck Davoine.
\newblock Event-based semantic-aided motion segmentation.
\newblock In \emph{Int. Conf. Comput. Vis. Theory Appl. (VISAPP)}, 2024.

\bibitem[Kim and Kim(2021)]{Kim21ral}
Haram Kim and H.~Jin Kim.
\newblock Real-time rotational motion estimation with contrast maximization
  over globally aligned events.
\newblock \emph{{IEEE} Robot. Autom. Lett.}, 6\penalty0 (3):\penalty0
  6016--6023, 2021.

\bibitem[Kingma and Ba(2015)]{Kingma15iclr}
Diederik~P. Kingma and Jimmy~L. Ba.
\newblock Adam: A method for stochastic optimization.
\newblock \emph{Int. Conf. Learn. Representations ({ICLR})}, 2015.

\bibitem[Lichtsteiner et~al.(2008)Lichtsteiner, Posch, and
  Delbruck]{Lichtsteiner08ssc}
Patrick Lichtsteiner, Christoph Posch, and Tobi Delbruck.
\newblock {A 128$\times$128 120 dB 15 $\mu$s latency asynchronous temporal
  contrast vision sensor}.
\newblock \emph{{IEEE} J. Solid-State Circuits}, 43\penalty0 (2):\penalty0
  566--576, 2008.

\bibitem[Mitrokhin et~al.(2018)Mitrokhin, Fermuller, Parameshwara, and
  Aloimonos]{Mitrokhin18iros}
Anton Mitrokhin, Cornelia Fermuller, Chethan Parameshwara, and Yiannis
  Aloimonos.
\newblock Event-based moving object detection and tracking.
\newblock In \emph{IEEE/RSJ Int. Conf. Intell. Robot. Syst. (IROS)}, pages
  1--9, 2018.

\bibitem[Mitrokhin et~al.(2019)Mitrokhin, Ye, Fermuller, Aloimonos, and
  Delbruck]{Mitrokhin19iros}
Anton Mitrokhin, Chengxi Ye, Cornelia Fermuller, Yiannis Aloimonos, and Tobi
  Delbruck.
\newblock {EV}-{IMO}: Motion segmentation dataset and learning pipeline for
  event cameras.
\newblock In \emph{IEEE/RSJ Int. Conf. Intell. Robot. Syst. (IROS)}, pages
  6105--6112, 2019.

\bibitem[Mitrokhin et~al.(2020)Mitrokhin, Hua, Ferm{\"u}ller, and
  Aloimonos]{Mitrokhin20cvpr}
Anton Mitrokhin, Zhiyuan Hua, Cornelia Ferm{\"u}ller, and Yiannis Aloimonos.
\newblock Learning visual motion segmentation using event surfaces.
\newblock In \emph{{IEEE} Conf. Comput. Vis. Pattern Recog. (CVPR)}, pages
  14402--14411, 2020.

\bibitem[Mueggler et~al.(2017)Mueggler, Rebecq, Gallego, Delbruck, and
  Scaramuzza]{Mueggler17ijrr}
Elias Mueggler, Henri Rebecq, Guillermo Gallego, Tobi Delbruck, and Davide
  Scaramuzza.
\newblock The event-camera dataset and simulator: Event-based data for pose
  estimation, visual odometry, and {SLAM}.
\newblock \emph{Int. J. Robot. Research}, 36\penalty0 (2):\penalty0 142--149,
  2017.

\bibitem[Mueggler et~al.(2018)Mueggler, Gallego, Rebecq, and
  Scaramuzza]{Mueggler18tro}
Elias Mueggler, Guillermo Gallego, Henri Rebecq, and Davide Scaramuzza.
\newblock Continuous-time visual-inertial odometry for event cameras.
\newblock \emph{{IEEE} Trans. Robot.}, 34\penalty0 (6):\penalty0 1425--1440,
  2018.

\bibitem[Ng et~al.(2022)Ng, Er, Soh, and Foong]{Ng22ral}
Matthew Ng, Zi~Min Er, Gim~Song Soh, and Shaohui Foong.
\newblock Aggregation functions for simultaneous attitude and image estimation
  with event cameras at high angular rates.
\newblock \emph{{IEEE} Robot. Autom. Lett.}, pages 1--1, 2022.

\bibitem[Nunes and Demiris(2021)]{Nunes21pami}
Urbano~Miguel Nunes and Yiannis Demiris.
\newblock Robust event-based vision model estimation by dispersion
  minimisation.
\newblock \emph{{IEEE} Trans. Pattern Anal. Mach. Intell.}, 2021.

\bibitem[Otsu(1975)]{Otsu75tsmc}
Nobuyuki Otsu.
\newblock A threshold selection method from gray-level histograms.
\newblock \emph{{IEEE} Trans. Systems, Man and Cybernetics}, 11\penalty0
  (285-296):\penalty0 23--27, 1975.

\bibitem[Parameshwara et~al.(2021{\natexlab{a}})Parameshwara, Li,
  Ferm{\"u}ller, Sanket, Evanusa, and Aloimonos]{Parameshwara21iros}
Chethan~M. Parameshwara, Simin Li, Cornelia Ferm{\"u}ller, Nitin~J. Sanket,
  Matthew~S. Evanusa, and Yiannis Aloimonos.
\newblock {SpikeMS}: Deep spiking neural network for motion segmentation.
\newblock In \emph{IEEE/RSJ Int. Conf. Intell. Robot. Syst. (IROS)}, pages
  3414--3420, 2021{\natexlab{a}}.

\bibitem[Parameshwara et~al.(2021{\natexlab{b}})Parameshwara, Sanket, Singh,
  Ferm{\"u}ller, and Aloimonos]{Parameshwara21icra}
Chethan~M. Parameshwara, Nitin~J. Sanket, Chahat~Deep Singh, Cornelia
  Ferm{\"u}ller, and Yiannis Aloimonos.
\newblock 0-{MMS}: Zero-shot multi-motion segmentation with a monocular event
  camera.
\newblock In \emph{{IEEE} Int. Conf. Robot. Autom. (ICRA)}, pages 9594--9600,
  2021{\natexlab{b}}.

\bibitem[Paredes-Vall^^c3^^a9s et~al.(2024)Paredes-Vall^^c3^^a9s, Hagenaars,
  Dupeyroux, Stroobants, Xu, and de~Croon]{Paredes24scirob}
Federico Paredes-Vall^^c3^^a9s, Jesse Hagenaars, Julien Dupeyroux, Stein
  Stroobants, Yingfu Xu, and Guido C. H.~E. de Croon.
\newblock Fully neuromorphic vision and control for autonomous drone flight.
\newblock \emph{Science Robotics}, 9\penalty0 (90):\penalty0 eadi0591, 2024.

\bibitem[Paredes-Vall{\'e}s et~al.(2023)Paredes-Vall{\'e}s, Scheper, De~Wagter,
  and de~Croon]{Paredes23iccv}
Federico Paredes-Vall{\'e}s, Kirk~YW Scheper, Christophe De~Wagter, and
  Guido~CHE de Croon.
\newblock Taming contrast maximization for learning sequential, low-latency,
  event-based optical flow.
\newblock In \emph{Int. Conf. Comput. Vis. (ICCV)}, pages 9661--9671, 2023.

\bibitem[Peng et~al.(2022)Peng, Gao, Wang, and Kneip]{Peng21pami}
Xin Peng, Ling Gao, Yifu Wang, and Laurent Kneip.
\newblock Globally-optimal contrast maximisation for event cameras.
\newblock \emph{{IEEE} Trans. Pattern Anal. Mach. Intell.}, 44\penalty0
  (7):\penalty0 3479--3495, 2022.

\bibitem[Posch et~al.(2014)Posch, Serrano-Gotarredona, Linares-Barranco, and
  Delbruck]{Posch14ieee}
Christoph Posch, Teresa Serrano-Gotarredona, Bernabe Linares-Barranco, and Tobi
  Delbruck.
\newblock Retinomorphic event-based vision sensors: Bioinspired cameras with
  spiking output.
\newblock \emph{Proc. {IEEE}}, 102\penalty0 (10):\penalty0 1470--1484, 2014.

\bibitem[{Rosinol Vidal} et~al.(2018){Rosinol Vidal}, Rebecq, Horstschaefer,
  and Scaramuzza]{Rosinol18ral}
Antoni {Rosinol Vidal}, Henri Rebecq, Timo Horstschaefer, and Davide
  Scaramuzza.
\newblock Ultimate {SLAM}? combining events, images, and {IMU} for robust
  visual {SLAM} in {HDR} and high speed scenarios.
\newblock \emph{{IEEE} Robot. Autom. Lett.}, 3\penalty0 (2):\penalty0
  994--1001, 2018.

\bibitem[Shiba et~al.(2022{\natexlab{a}})Shiba, Aoki, and Gallego]{Shiba22aisy}
Shintaro Shiba, Yoshimitsu Aoki, and Guillermo Gallego.
\newblock A fast geometric regularizer to mitigate event collapse in the
  contrast maximization framework.
\newblock \emph{Adv. Intell. Syst.}, page 2200251, 2022{\natexlab{a}}.

\bibitem[Shiba et~al.(2022{\natexlab{b}})Shiba, Aoki, and Gallego]{Shiba22eccv}
Shintaro Shiba, Yoshimitsu Aoki, and Guillermo Gallego.
\newblock Secrets of event-based optical flow.
\newblock In \emph{Eur. Conf. Comput. Vis. (ECCV)}, pages 628--645,
  2022{\natexlab{b}}.

\bibitem[Shiba et~al.(2022{\natexlab{c}})Shiba, Aoki, and
  Gallego]{Shiba22sensors}
Shintaro Shiba, Yoshimitsu Aoki, and Guillermo Gallego.
\newblock Event collapse in contrast maximization frameworks.
\newblock \emph{Sensors}, 22\penalty0 (14):\penalty0 1--20, 2022{\natexlab{c}}.

\bibitem[Shiba et~al.(2024)Shiba, Klose, Aoki, and Gallego]{Shiba24pami}
Shintaro Shiba, Yannick Klose, Yoshimitsu Aoki, and Guillermo Gallego.
\newblock Secrets of event-based optical flow, depth, and ego-motion by
  contrast maximization.
\newblock \emph{{IEEE} Trans. Pattern Anal. Mach. Intell.}, 46\penalty0
  (12):\penalty0 7742--7759, 2024.

\bibitem[Stoffregen and Kleeman(2019)]{Stoffregen19cvpr}
Timo Stoffregen and Lindsay Kleeman.
\newblock Event cameras, contrast maximization and reward functions: an
  analysis.
\newblock In \emph{{IEEE} Conf. Comput. Vis. Pattern Recog. (CVPR)}, pages
  12292--12300, 2019.

\bibitem[Stoffregen et~al.(2019)Stoffregen, Gallego, Drummond, Kleeman, and
  Scaramuzza]{Stoffregen19iccv}
Timo Stoffregen, Guillermo Gallego, Tom Drummond, Lindsay Kleeman, and Davide
  Scaramuzza.
\newblock Event-based motion segmentation by motion compensation.
\newblock In \emph{Int. Conf. Comput. Vis. (ICCV)}, pages 7243--7252, 2019.

\bibitem[Stoffregen et~al.(2020)Stoffregen, Scheerlinck, Scaramuzza, Drummond,
  Barnes, Kleeman, and Mahony]{Stoffregen20eccv}
Timo Stoffregen, Cedric Scheerlinck, Davide Scaramuzza, Tom Drummond, Nick
  Barnes, Lindsay Kleeman, and Robert Mahony.
\newblock Reducing the sim-to-real gap for event cameras.
\newblock In \emph{Eur. Conf. Comput. Vis. (ECCV)}, pages 534--549, 2020.

\bibitem[Taverni et~al.(2018)Taverni, Moeys, Li, Cavaco, Motsnyi, Bello, and
  Delbruck]{Taverni18tcsii}
Gemma Taverni, Diederik~Paul Moeys, Chenghan Li, Celso Cavaco, Vasyl Motsnyi,
  David San~Segundo Bello, and Tobi Delbruck.
\newblock Front and back illuminated {D}ynamic and {A}ctive {P}ixel {V}ision
  {S}ensors comparison.
\newblock \emph{{IEEE} Trans. Circuits Syst. {II}}, 65\penalty0 (5):\penalty0
  677--681, 2018.

\bibitem[Trucco and Verri(1998)]{Trucco98book}
Emanuele Trucco and Alessandro Verri.
\newblock \emph{Introductory Techniques for {3-D} Computer Vision}.
\newblock Prentice Hall PTR, Upper Saddle River, NJ, USA, 1998.

\bibitem[Wang et~al.(2024)Wang, Guo, and Daniilidis]{Wang24evmoseg}
Ziyun Wang, Jinyuan Guo, and Kostas Daniilidis.
\newblock {Un-EVIMO}: Unsupervised event-based independent motion segmentation.
\newblock In \emph{Eur. Conf. Comput. Vis. (ECCV)}, 2024.

\bibitem[Zhao et~al.(2023)Zhao, Li, and Lyu]{zhao23icra}
Chunhui Zhao, Yakun Li, and Yang Lyu.
\newblock Event-based real-time moving object detection based on imu ego-motion
  compensation.
\newblock In \emph{{IEEE} Int. Conf. Robot. Autom. (ICRA)}, pages 690--696,
  2023.

\bibitem[Zhou et~al.(2024)Zhou, Shi, Dong, Peng, Chang, and Yan]{zhou2024jstr}
Hanyu Zhou, Zhiwei Shi, Hao Dong, Shihan Peng, Yi Chang, and Luxin Yan.
\newblock {JSTR}: Joint spatio-temporal reasoning for event-based moving object
  detection.
\newblock In \emph{{IEEE} Int. Conf. Robot. Autom. (ICRA)}, pages 10650--10656,
  2024.

\bibitem[Zhou et~al.(2021)Zhou, Gallego, Lu, Liu, and Shen]{Zhou21tnnls}
Yi Zhou, Guillermo Gallego, Xiuyuan Lu, Siqi Liu, and Shaojie Shen.
\newblock Event-based motion segmentation with spatio-temporal graph cuts.
\newblock \emph{{IEEE} Trans. Neural Netw. Learn. Syst.}, pages 1--13, 2021.

\bibitem[Zhu et~al.(2017)Zhu, Atanasov, and Daniilidis]{Zhu17cvpr}
Alex~Zihao Zhu, Nikolay Atanasov, and Kostas Daniilidis.
\newblock Event-based visual inertial odometry.
\newblock In \emph{{IEEE} Conf. Comput. Vis. Pattern Recog. (CVPR)}, pages
  5816--5824, 2017.

\end{thebibliography}
}

\cleardoublepage

\fi

\end{document}